\documentclass{article}

\usepackage{microtype}
\usepackage{graphicx}
\usepackage{subfigure}
\usepackage{booktabs} 
\usepackage{enumitem}
\usepackage{tcolorbox}
\usepackage{wrapfig}
\tcbuselibrary{breakable}
\usepackage{multirow}
\usepackage{diagbox}
\usepackage{subfigure}
\usepackage{pifont}
\usepackage{hyperref}
\NewDocumentCommand{\cdp}
{ mO{} }{\textcolor{blue}{\textsuperscript{\textit{Dong ping}}\textsf{\textbf{\small[#1]}}}}
\NewDocumentCommand{\yinuo}
{ mO{} }{\textcolor{blue}{\textsuperscript{\textit{Yi Nuo}}\textsf{\textbf{\small[#1]}}}}

\usepackage[accepted]{icml2024}

\usepackage{amsmath}
\usepackage{amssymb}
\usepackage{mathtools}
\usepackage{amsthm}

\usepackage[capitalize,noabbrev]{cleveref}

\theoremstyle{plain}

\theoremstyle{definition}

\theoremstyle{remark}

\usepackage[textsize=tiny]{todonotes}

\icmltitlerunning{MLLM-as-a-Judge: Assessing Multimodal LLM-as-a-Judge with Vision-Language Benchmark}

\begin{document}

\twocolumn[
\icmltitle{MLLM-as-a-Judge:\\ Assessing Multimodal LLM-as-a-Judge with Vision-Language Benchmark
}



\icmlsetsymbol{equal}{*}

\begin{icmlauthorlist}
\icmlauthor{Dongping Chen}{equal,hust}
\icmlauthor{Ruoxi Chen}{equal,zjut}
\icmlauthor{Shilin Zhang}{equal,hust}
\icmlauthor{Yaochen Wang}{equal,hust}
\icmlauthor{Yinuo Liu}{equal,hust}
\icmlauthor{Huichi Zhou}{equal,hust}
\icmlauthor{Qihui Zhang}{equal,hust}
\icmlauthor{Yao Wan}{hust}
\icmlauthor{Pan Zhou}{hust}
\icmlauthor{Lichao Sun}{lehigh}
\end{icmlauthorlist}

\icmlaffiliation{hust}{Huazhong University of Science and Technology}
\icmlaffiliation{zjut}{Zhejiang University of Technology}
\icmlaffiliation{lehigh}{LAIR Lab, Lehigh University}

\icmlcorrespondingauthor{Yao Wan}{wanyao@hust.edu.cn}
\icmlcorrespondingauthor{Pan Zhou}{panzhou@hust.edu.cn}

\vskip 0.3in
]



\printAffiliationsAndNotice{\icmlEqualContribution} 

\begin{abstract}
Multimodal Large Language Models (MLLMs) have gained significant attention recently, showing remarkable potential in artificial general intelligence.
However, assessing the utility of MLLMs presents considerable challenges, primarily due to the absence of multimodal benchmarks that align with human preferences.
Drawing inspiration from the concept of LLM-as-a-Judge within LLMs, this paper introduces a novel benchmark, termed \text{MLLM-as-a-Judge}, to assess the ability of MLLMs in assisting judges across diverse modalities, encompassing three distinct tasks: \textit{Scoring Evaluation}, \textit{Pair Comparison}, and \textit{Batch Ranking}.
Our study reveals that, while MLLMs demonstrate remarkable human-like discernment in \textit{Pair Comparison}, there is a significant divergence from human preferences in \textit{Scoring Evaluation} and \textit{Batch Ranking}.
Furthermore, a closer examination reveals persistent challenges in the judgment capacities of LLMs, including diverse biases, hallucinatory responses, and inconsistencies in judgment, even in advanced models such as GPT-4V.
These findings emphasize the pressing need for enhancements and further research efforts to be undertaken before regarding MLLMs as fully reliable evaluators. 
In light of this, we advocate for additional efforts dedicated to supporting the continuous development within the domain of MLLM functioning as judges. The code and dataset are publicly available at our project homepage: \url{https://mllm-judge.github.io/}.

\end{abstract}
\section{Introduction}
The advent of Large Language Models (LLMs), such as GPT-3~\cite{openai2023gpt4} and Llama~\cite{touvron2023llama}, has achieved substantial progress in content generation, 
including text generation~\citep{openai2023gpt4}, code generation~\cite{roziere2023code}, and video synthesis~\citep{wu2023next}. The emergent abilities of LLMs, as demonstrated by the Chain-of-Thought (CoT) framework~\cite{wei2022chain}, present a promising avenue for their utilization as evaluators, also referred to as the LLM-as-a-Judge~\citep{zheng2023judging}. Initial explorations indicate a better alignment with human preferences, emphasizing the considerable potential inherent in this approach.

Recently, building upon LLMs, Multimodal Large Language Models (MLLMs) like GPT-4V~\cite{openai2023gpt4v} and LLaVA~\cite{liu2023llava} exhibit exceptional proficiency by incorporating multiple modalities (e.g., text, charts, images, and videos) and showcasing remarkable performance in multimodal applications, including text-to-video~\cite{wu2023next} and visual dialog~\cite{cai2023low}.
Despite this, 
assessing the effectiveness of MLLMs remains challenging due to the limitations of traditional metrics, which hinge on text-based exact matches or embedding distances. These metrics fall short in adhering to the granular evaluation criteria of interest and fail to capture the rich context within the generated outputs.
Drawing inspiration from the concept of LLM-as-a-Judge within LLMs, a pertinent research question arises: \textit{``Can MLLMs effectively serve as judges in the multimodal domain, and how closely do their evaluations align with human preferences?''}

\begin{figure*}[t]
\vspace{-5pt}
    \centering
    \includegraphics[width=.88\linewidth]{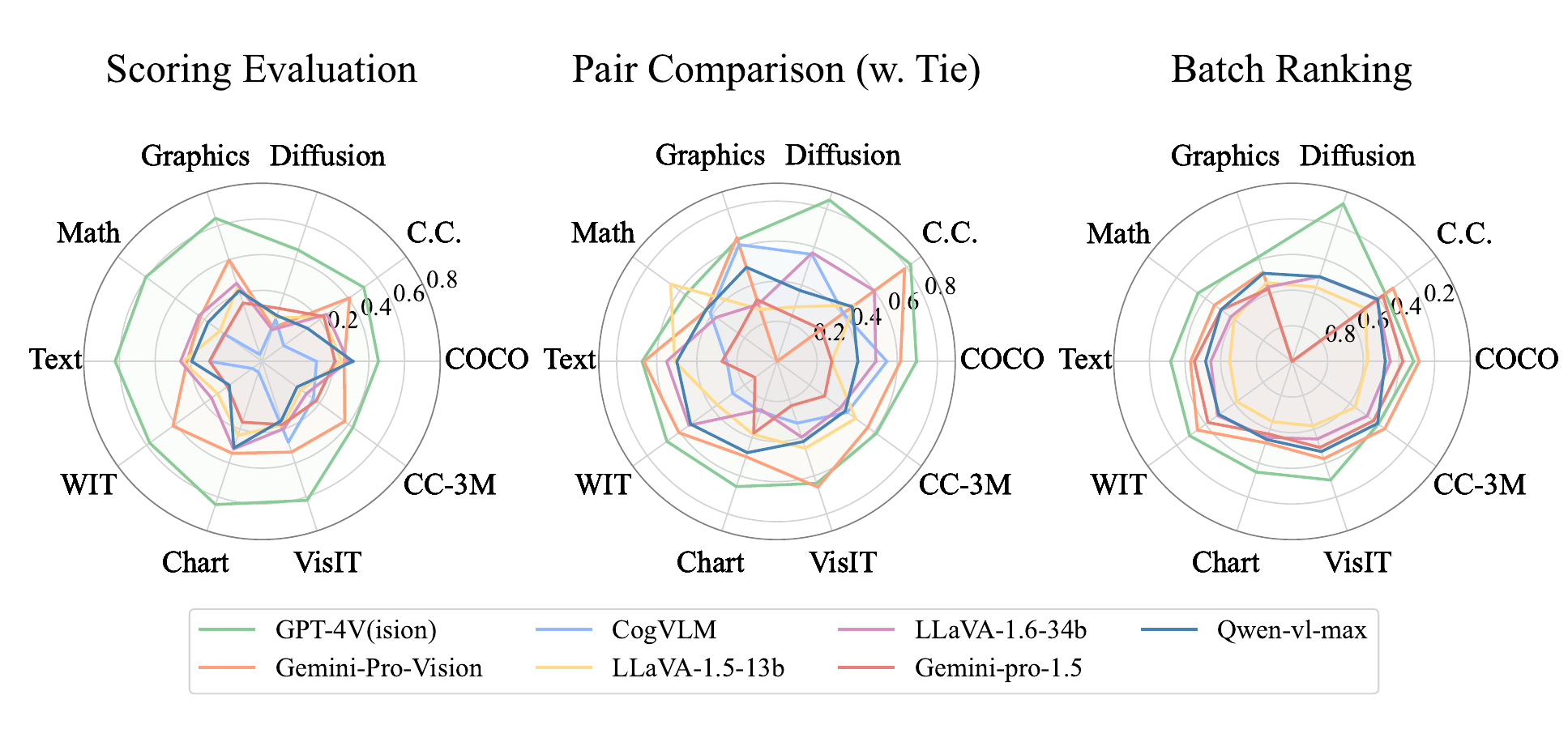}
      \vspace{-13pt}
    \caption{Comparative performance of different MLLMs across three judging settings in 10 datasets, each is the average of three iterations. As the CogVLM is unable to perform the batch ranking task, we show the other six MLLMs only.}
    \label{fig: Radar Figure}
      \vspace{-12pt}
\end{figure*}

To answer this question, this paper undertakes an extensive study, introducing a groundbreaking benchmark, MLLM-as-a-Judge, specifically crafted to evaluate the efficacy of MLLMs in assisting judges across diverse modalities.
To achieve this goal, we first thoughtfully curate a selection of 14 datasets across various tasks, including image captioning, math reasoning, text reading, and infographics understanding, culminating in acquiring a dataset comprising 4,414 image-instruction pairs.
Subsequently, we utilize six main-stream MLLMs from a model pool which includes GPT-4V \citep{openai2023gpt4v}, Gemini \citep{geminiteam2023gemini}\footnote{For conciseness, we refer to  GPT-4V(ision) as GPT-4V, and Gemini-Pro-Vision as Gemini throughout this paper.}, LLaVA-1.5-13b, LLaVA-1.6-34b \citep{liu2023llava}, CogVLM \citep{wang2023cogvlm}, Qwen-VL-Max \citep{Qwen-VL}, to generate responses to each instruction across three distinct evaluation settings.
The produced responses are subsequently gathered and undergo additional annotation by human evaluators, who apply stringent criteria to ensure an impartial and thorough assessment of the judgments made by the MLLMs.

Furthermore, we assess the ability of MLLMs as judges in multimodal tasks by calculating the similarity between human and MLLMs judgment and measuring human agreement on the analysis and judgment made by those MLLMs.
In particular, we target eleven widely-used MLLMs, i.e., GPT-4V and Gemini-Pro-1.0/1.5, CogVLM, LLaVA-1.5/1.6 family, and Qwen-VL family, across two settings (with, or without vision input), over three distinct tasks (i.e., \textit{Scoring Evaluation}, \textit{Pair Comparison}, and \textit{Batch Ranking}). Figure~\ref{fig: Radar Figure} compares the performance of various MLLMs across different datasets and settings, illustrating that GPT-4V exhibits significantly superior capabilities as a judge compared to other MLLMs.

As a benchmark, we also release two curated datasets to facilitate further studies: \textsc{MLLM-as-a-Judge-HQ}, which showcases responses with a high level of concordance with human judgments, and \textsc{MLLM-as-a-Judge-Hard}, which includes responses marked by inconsistency with human preferences and instances of hallucination. Additionally, we address the limitations of MLLMs in judgment, such as egocentric bias, position bias, length bias, and hallucination. We demonstrate that integrating CoT \citep{wei2022chain} and a vision expert system can effectively mitigate some of these biases.

\paragraph{Take-Aways.}
We evaluate the judgment performance of 11 MLLMs across 14 datasets under three settings: score evaluation, pair comparison, and batch ranking. Our findings reveal several key insights. First, while MLLMs demonstrate proficiency in aligning with human preferences in pair comparison tasks, they require further improvement in score evaluation and batch ranking, particularly in reasoning tasks. Secondly, GPT-4V consistently outperforms other models across all tasks and settings.

Finally, the presence of hallucinations, biases, and inconsistent judgments in MLLMs highlights significant challenges that must be addressed for these models to become a viable alternative to traditional human evaluations.

To summarize, our work provides three key contributions: 
\vspace{-5pt}
\begin{itemize}[nolistsep, leftmargin=*]
    \item \textbf{A Benchmark.}
    We are the first to develop a comprehensive benchmark MLLM-as-a-Judge in multimodal domains, with human annotations to assess the judging capability of MLLMs in tasks of \textit{Scoring Evaluation}, \textit{Pair Comparison} and \textit{Batch Ranking}. 
    \item \textbf{Two Datasets.}
    We curate two human preference datasets: \textsc{MLLM-as-a-Judge-HQ}, which contains high-quality questions, and \textsc{MLLM-as-a-Judge-HARD}, which includes instances of hallucination. These datasets can serve as rigorous testing grounds to facilitate the development of MLLMs in aligning human preferences.

    \item \textbf{Findings and Implications.}
    Our evaluation of mainstream MLLMs reveals that while MLLMs exhibit alignment with human judgments in \textit{Pair Comparison}, notable discrepancies can be found in \textit{Scoring Evaluation} and \textit{Batch Ranking}. Furthermore, our findings reveal that MLLMs exhibit a range of biases and hallucinations, along with inconsistent judgments during the evaluation process, representing significant hurdles in establishing MLLMs as reliable judges.

\end{itemize}

\begin{figure*}[t]
    \centering
    \includegraphics[width=0.95\linewidth]{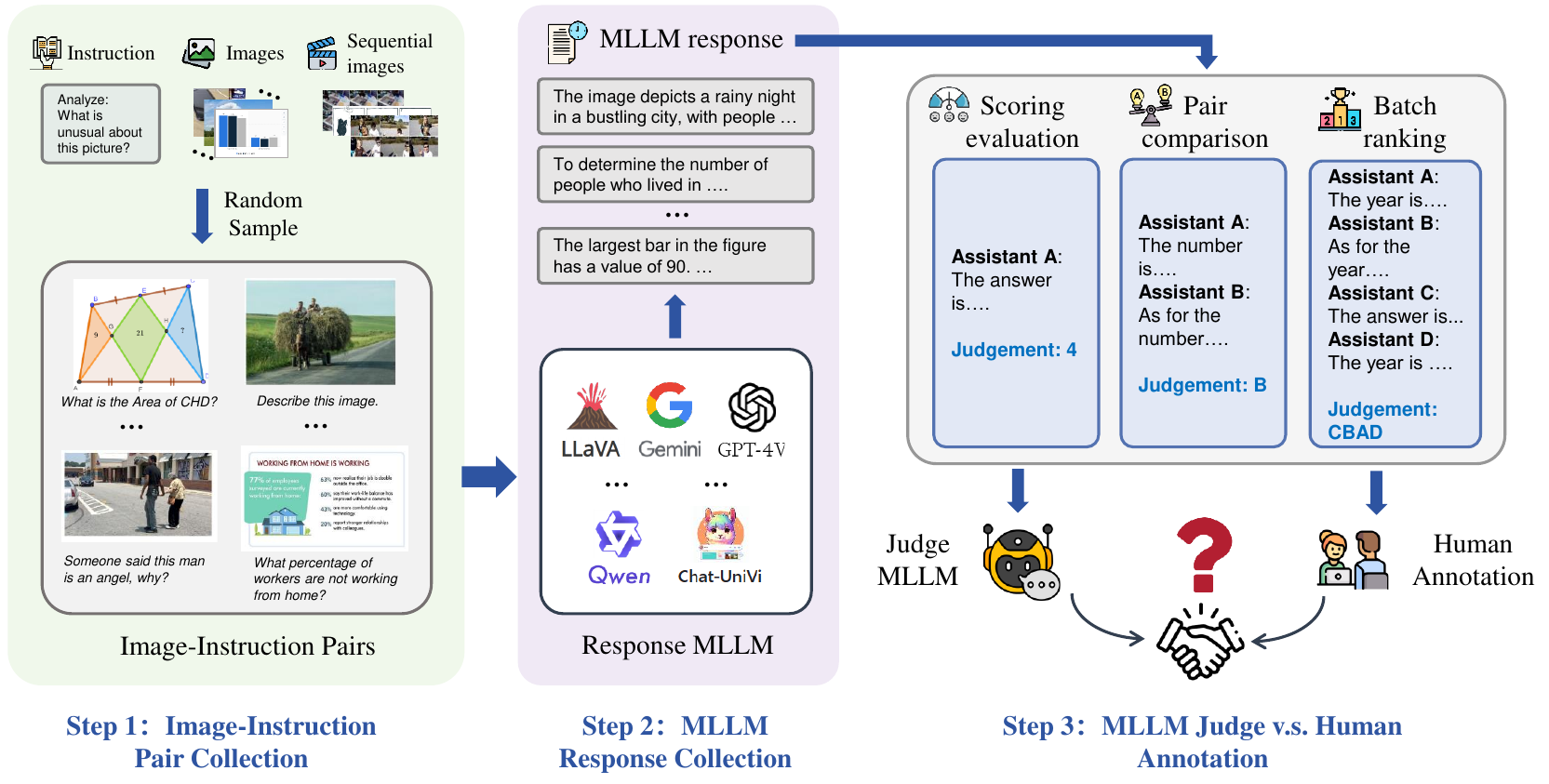}
    \vspace{-8pt}
    \caption{An overview of {MLLM-as-a-Judge}.}
      \vspace{-8pt}
    \label{fig: pipeline}
\end{figure*}

\section{{MLLM-as-a-Judge}: A Benchmark to Assess Vision-Language Judging Ability}
\label{Section 3: MLLM-as-a-Judge}

\begin{table}[t]
\centering
\caption{The statistics of responses in different steps for MLLM judging. In \textit{Step 3}, under the w.o. vision input settings, we sample 10\% from the original data and mainly proceed with \text{GPT-4V} and \text{Gemini}. We only list the amount of judgments generated by four models here. M-I: Image-Instruction.}
\renewcommand\arraystretch{1.1}
\label{table: comparison}
\resizebox{0.98\linewidth}{!}{
\begin{tabular}{ccclclc}
\toprule[1.5pt]
\textbf{Step}           & \multicolumn{2}{c}{\textbf{Setting}}                                   & \textbf{Input}                                                                     & \textbf{Num.}          & \textbf{Output}                                                                    & \textbf{Num.}          \\ \midrule
\multirow{2}{*}{1}  & \multicolumn{2}{c}{\multirow{2}{*}{/}}                                 & Image                                                                              & 4,144                  & \multirow{2}{*}{\begin{tabular}[c]{@{}l@{}}M-I Pairs\end{tabular}} & \multirow{2}{*}{4,400} \\
                        & \multicolumn{2}{c}{}                                                   & Instruction                                                                        & 4,414                  &                                                                                    &                       \\ \midrule
2  & \multicolumn{2}{c}{/}                                 & \begin{tabular}[c]{@{}l@{}} M-I Pairs\end{tabular} & 3,300 & MLLMs & 17,096\\ \midrule
\multirow{25}{*}{3} & \multicolumn{2}{c}{\multirow{12}{*}{\rotatebox{90}{w. Vision Input} }}                   & \multirow{4}{*}{Batch}                                                             & \multirow{4}{*}{1,470} & \text{Gemini}                                                                             & 1,340                   \\
                        & \multicolumn{2}{c}{}                                                   &                                                                                    &                       & \text{GPT-4V}                                                                                & 1,454                \\
                        & \multicolumn{2}{c}{}                                                   &                                                                                    &                       & \text{Qwen-VL-Max}                                                                             & 1,458                  \\
                         & \multicolumn{2}{c}{}                                                   &                                                                                    &                       & \text{LLaVA}                                                                             & 1,468                  \\\cline{4-7} 
                        & \multicolumn{2}{c}{}                                                   & \multirow{4}{*}{Pair}                                                              & \multirow{4}{*}{8,256} & \text{Gemini}                                                                             & 7,751                  \\
                        & \multicolumn{2}{c}{}                                                   &                                                                                    &                       & \text{GPT-4V}                                                                               & 8,117                  \\
                        & \multicolumn{2}{c}{}                                                   &                                                                                    &                       & \text{Qwen-VL-Max}                                                                             & 8,012                  \\ 
                        & \multicolumn{2}{c}{}                                                   &                                                                                    &                       & \text{LLaVA}                                                                             & 8,253                  \\ 
                        \cline{4-7} 
                        & \multicolumn{2}{c}{}                                                   & \multirow{4}{*}{Score}                                                             & \multirow{4}{*}{5,883} & \text{Gemini}                                                                             & 5,337                  \\
                        & \multicolumn{2}{c}{}                                                   &                                                                                    &                       & \text{GPT-4V}                                                                                & 5,708                  \\
                        & \multicolumn{2}{c}{}                                                   &                                                                                    &                       & \text{Qwen-VL-Max}                                                                             & 5,701                   \\
                        
                        & \multicolumn{2}{c}{}                                                   &                                                                                    &                       & \text{LLaVA}                                                                             &  5,729              \\
                        \cline{2-7} 
                        & \multirow{12}{*}{\rotatebox{90}{w.o. Vision Input} } & \multirow{6}{*}{\rotatebox{90}{No Vision} } & \multirow{2}{*}{Batch}                                                             & \multirow{2}{*}{110}  & \text{Gemini}                                                                             & 107                   \\
                        &                                    &                                   &                                                                                    &                       & \text{GPT-4V}                                                                                & 110                   \\ \cline{4-7} 
                        &                                    &                                   & \multirow{2}{*}{Pair}                                                              & \multirow{2}{*}{425}  & \text{Gemini}                                                                             & 385                   \\
                        &                                    &                                   &                                                                                    &                       & \text{GPT-4V}                                                                                & 355                   \\ \cline{4-7} 
                        &                                    &                                   & \multirow{2}{*}{Score}                                                             & \multirow{2}{*}{612}  & \text{Gemini}                                                                             & 582                   \\
                        &                                    &                                   &                                                                                    &                       & \text{GPT-4V}                                                                                & 584                   \\ \cline{3-7} 
                        &                                    & \multirow{6}{*}{\rotatebox{90}{Vision Experts}}   & \multirow{2}{*}{Batch}                                                             & \multirow{2}{*}{110}  & \text{Gemini}                                                                             & 107                   \\
                        &                                    &                                   &                                                                                    &                       & \text{GPT-4V}                                                                                & 110                   \\ \cline{4-7} 
                        &                                    &                                   & \multirow{2}{*}{Pair}                                                              & \multirow{2}{*}{425}  & \text{Gemini}                                                                             & 396                   \\
                        &                                    &                                   &                                                                                    &                       & \text{GPT-4V}                                                                                & 425                   \\ \cline{4-7} 
                        &                                    &                                   & \multirow{2}{*}{Score}                                                             & \multirow{2}{*}{612}  & \text{Gemini}                                                                             & 576                   \\
                        &                                    &                                   &                                                                                    &                       & \text{GPT-4V}                                                                                & 612                   \\ \bottomrule[1.5pt]
\end{tabular}}
  \vspace{-15pt}
\end{table}

Figure \ref{fig: pipeline} shows an overview of our proposed MLLM-as-a-Judge, consisting of three steps: 1) image-instruction pair collection, 2) MLLM response collection, and 3) comparison with human annotation.
Initially, we collect a dataset $\mathcal{P} = \{(M_1, I_1), \ldots, (M_n, I_n)\}$, containing pairs of images $(M)$ and their corresponding instructions $(I)$ sourced from 10 diverse domains (e.g., math, chart, diffusion), ensuring comprehensive coverage for a wide array of downstream tasks.
Subsequently, each pair $(M_i, I_i)$ is processed through several MLLMs, 
generating a set of responses $\mathcal{R}_i = \{r_1, r_2, \ldots, r_n\}$ for each pair. This process contributes to the formation of the dataset of image-instruction-responses pairs, denoted as $\mathcal{D} = \{(M_i, I_i, \mathcal{R}_i) | (M_i, I_i) \in \mathcal{P}\}$.
Finally, the dataset $\mathcal{D}$ is partitioned into three distinct subsets to facilitate diverse task evaluations: $\mathcal{D}_{\text{score}}$ for \textit{Scoring Evaluation}, $\mathcal{D}_{\text{pair}}$ for \textit{Pair Comparison}, and $\mathcal{D}_{\text{batch}}$ for \textit{Batch Ranking}.
Each subset will be employed for specific judging tasks, with each of them being configured as follows.
\begin{itemize}[nolistsep, leftmargin=*]
\item \textbf{Scoring Evaluation}: Each individual response is evaluated on a scale from 1 to 5, with the specific criteria for this rating system detailed in Appendix~\ref{Prompt templates}.
\item \textbf{Pair Comparison}: It involves a direct comparison between two responses, 
culminating in the identification of the superior one. Following the principles outlined by \citep{deutsch2023ties}, a tie option is incorporated to ensure a more equitable assessment.
\item \textbf{Batch Ranking}: 
The responses are systematically arranged in descending order of quality based on a given instruction, without any tie option.
\end{itemize} 

\subsection{Step 1: Image-Instruction Pair Collection}
We meticulously curate a dataset consisting of 4,414 image-text pairs, gathered from a variety of downstream task datasets, as detailed in Table~\ref{Step1: Detailed Dataset} in Appendix~\ref{Detailed Benchmark Construction}. These pairs are carefully tailored into image-instruction pairs to suit a free-form response format. To illustrate, within the domain of diffusion tasks, our dataset incorporated pairs challenging models to adeptly recognize and articulate connections between provided images and user-specified keywords.

\subsection{Step 2: MLLM Response Collection}

We employ six widely-used MLLMs – \text{GPT-4V} \citep{openai2023gpt4v}, \text{Gemini} \citep{geminiteam2023gemini}, \text{LLaVA} \citep{liu2023llava}, Qwen-VL-Max \citep{Qwen-VL}, LLaVA-1.6-34b \citep{liu2023llava}, and \text{CogVLM} \citep{wang2023cogvlm} – to generate responses based on the image-instruction pairs, obtaining approximately 17,000 responses. 
Responses that are either too brief or non-compliant with security regulations (e.g., \textit{``I'm sorry, but I cannot assist with this request''}) from \text{GPT-4V} and \text{Gemini} are excluded. 
The number of responses and the length distributions for different MLLMs are shown in Table~\ref{table: comparison} and Figure~\ref{fig:length_distribution}, respectively. 
We show specific hyper-parameter settings in Appendix~\ref{Detailed Benchmark Construction: Step 2}. Besides, we segment these responses into three non-overlapping groups, to prevent response overlap. 

\begin{figure}[t]
    \centering
    \includegraphics[width=0.8\linewidth]{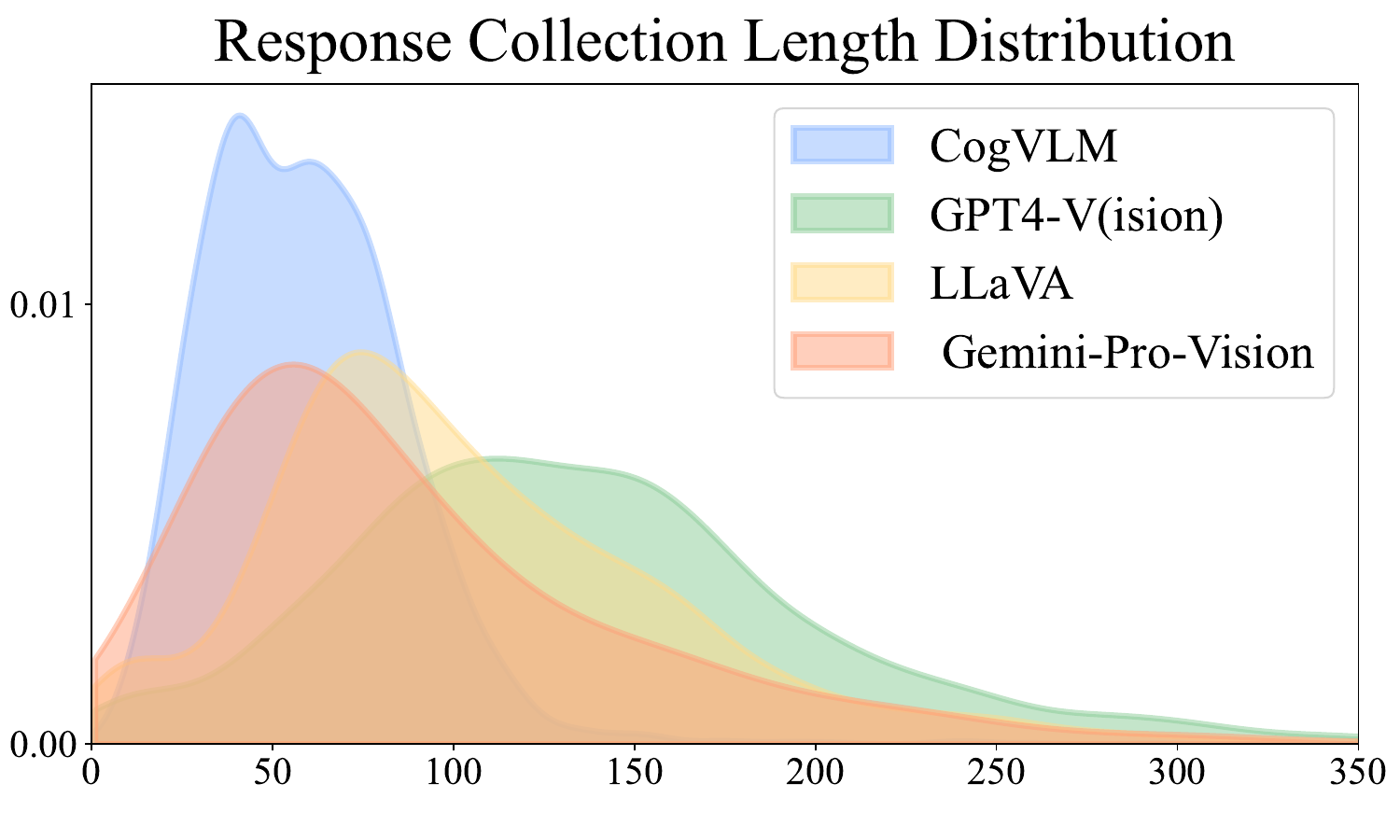}
    \vspace{-15pt}
    \caption{Length distribution in responses for different MLLMs. Horizontal axis: length; Vertical axis: density.}
    \label{fig:length_distribution}
    \vspace{-10pt}
\end{figure}

\subsection{Step 3: Comparison with Human Annotations}
The annotation is conducted by 6 authors of this paper independently. These annotators are proficient in this domain, with different genders, ages, and educational backgrounds to ensure diversity~\citep{sun2020evolution}. They are required to give objective judgments without considering answer lengths, and certain names or positions of the response to minimize human bias. More details are referred to Appendix~\ref{Human Labeling and Agreement Collection}.

\begin{table*}[htbp]
\centering
\large
\renewcommand\arraystretch{1.2}
\label{tab: experiment result}
\caption{The overall performance of different MLLMs in judging, compared with human annotations on different datasets. We sample all the data three times and took the average to mitigate the casualty. \textit{w.} and \textit{w.o.} tie represents tie and non-tie situations respectively. We omit Gemini's results on the diffusion task for its challenges in processing AI-generated images. 
All presented data of Pearson similarity exhibit a $p$-value below 0.05, indicating a statistically significant level of confidence. Please refer to the Appendix~\ref{full} for more results.
}
\resizebox{1\linewidth}{!}{
\begin{tabular}{ll|ccccccccccccccc}\toprule[1.5pt]
\textbf{Settings} & \textbf{MLLM}  & COCO  & C.C. & Diff. & Graphics & Math & Text & WIT & Chart & VisIT & CC-3M & M2W & SciQA & Aes & MM-Vet & Ave. \\ \midrule
\multirow{5}{*}{\textbf{Score ($\uparrow$)}} & LLaVA-1.5-13b & 0.247 & 0.227 & 0.060 & 0.242 & 0.093 & 0.245 & 0.109 & 0.237 & 0.177 & 0.071 & \textbf{0.424} & 0.279 & \textbf{0.414} & 0.322 & 0.225 \\
 & LLaVA-1.6-34b & 0.285 & 0.251 & -0.012 & 0.262 & 0.238 & 0.258 & 0.151 & 0.318 & 0.198 & 0.109 & 0.022 & 0.206 & 0.025 & 0.265 & 0.184 \\
 & Gemini & 0.262 & 0.408 & - & 0.400 & 0.228 & 0.222 & 0.418 & 0.343 & 0.336 & 0.374 & 0.324 & 0.073 & 0.360 & 0.207 & 0.304 \\
 & GPT-4V & \textbf{0.454} & \textbf{0.507} & \textbf{0.458} & \textbf{0.645} & \textbf{0.606} & \textbf{0.624} & \textbf{0.579} & \textbf{0.645} & \textbf{0.620} & \textbf{0.431} & 0.185 & \textbf{0.383} & 0.401 & \textbf{0.326} & \textbf{0.490} \\
 & Qwen-vl-max & 0.311 & 0.117 & 0.072 & 0.218 & 0.175 & 0.196 & 0.028 & 0.312 & 0.151 & 0.045 & 0.244 & 0.115 & 0.177 & 0.216 & 0.170 \\ \midrule
\multirow{5}{*}{\textbf{Pair w. Tie ($\uparrow$)}} & LLaVA-1.5-13b & 0.273 & 0.478 & 0.286 & 0.273 & \textbf{0.657} & 0.510 & 0.369 & 0.383 & 0.456 & 0.484 & 0.347 & 0.223 & 0.389 & 0.254 & 0.384 \\
 & LLaVA-1.6-34b & 0.493 & 0.600 & 0.570 & 0.300 & 0.374 & 0.551 & 0.543 & 0.254 & 0.398 & 0.392 & 0.513 & \textbf{0.434} & 0.524 & 0.499 & 0.460 \\
 & Gemini & 0.616 & 0.787 & - & \textbf{0.650} & 0.436 & 0.664 & 0.605 & 0.500 & \textbf{0.660} & 0.560 & 0.370 & 0.262 & 0.190 & 0.312 & 0.509 \\
 & GPT-4V & \textbf{0.696} & \textbf{0.824} & \textbf{0.847} & 0.639 & 0.564 & \textbf{0.673} & \textbf{0.679} & \textbf{0.657} & 0.640 & \textbf{0.612} & \textbf{0.521} & 0.415 & \textbf{0.606} & \textbf{0.529} & \textbf{0.636} \\
 & Qwen-vl-max & 0.403 & 0.464 & 0.372 & 0.494 & 0.438 & 0.500 & 0.533 & 0.479 & 0.421 & 0.421 & 0.411 & 0.392 & 0.325 & 0.474 & 0.438 \\ \midrule
\multirow{5}{*}{\textbf{Pair w.o. Tie ($\uparrow$)}} & LLaVA-1.5-13b & 0.327 & 0.537 & 0.302 & 0.300 & 0.726 & 0.684 & 0.600 & 0.610 & 0.648 & 0.583 & 0.449 & 0.443 & 0.498 & 0.344 & 0.504 \\
 & LLaVA-1.6-34b & 0.607 & 0.824 & 0.855 & 0.402 & 0.587 & 0.750 & \textbf{0.758} & 0.381 & 0.503 & 0.564 & \textbf{0.712} & \textbf{0.679} & 0.694 & \textbf{0.762} & 0.648 \\
 & Gemini & 0.717 & 0.840 & - & 0.770 & 0.678 & 0.793 & 0.688 & 0.658 & 0.711 & 0.652 & 0.471 & 0.358 & 0.265 & 0.400 & 0.615 \\
 & GPT-4V & \textbf{0.804} & \textbf{0.870} & \textbf{0.922} & \textbf{0.807} & \textbf{0.801} & \textbf{0.805} & 0.734 & \textbf{0.849} & \textbf{0.761} & \textbf{0.703} & 0.699 & 0.647 & \textbf{0.755} & 0.659 & \textbf{0.773} \\
 & Qwen-vl-max & 0.657 & 0.674 & 0.556 & 0.667 & 0.635 & 0.732 & 0.647 & 0.638 & 0.560 & 0.586 & 0.608 & 0.646 & 0.741 & 0.662 & 0.644 \\ \midrule
\multirow{5}{*}{\textbf{Batch ($\downarrow$)}} & LLaVA-1.5-13b & 0.577 & 0.492 & 0.562 & 0.535 & 0.598 & 0.650 & 0.616 & 0.644 & 0.620 & 0.563 & 0.639 & 0.563 & 0.650 & 0.652 & 0.597 \\
 & LLaVA-1.6-34b & 0.449 & 0.411 & 0.500 & 0.561 & 0.575 & 0.544 & 0.483 & 0.552 & 0.542 & 0.479 & \textbf{0.529} & 0.437 & 0.500 & 0.450 & 0.501 \\
 & Gemini & \textbf{0.287} & \textbf{0.299} & - & 0.473 & 0.462 & 0.430 & 0.344 & 0.520 & 0.426 & 0.357 & 0.613 & \textbf{0.412} & 0.467 & 0.529 & 0.432 \\
 & GPT-4V & 0.318 & 0.353 & \textbf{0.070} & \textbf{0.385} & \textbf{0.348} & \textbf{0.319} & \textbf{0.290} & \textbf{0.347} & \textbf{0.300} & \textbf{0.402} & 0.597 & 0.462 & 0.453 & \textbf{0.411} & \textbf{0.361} \\
 & Qwen-vl-max & 0.477 & 0.407 & 0.500 & 0.480 & 0.507 & 0.515 & 0.493 & 0.539 & 0.468 & 0.407 & 0.563 & 0.503 & \textbf{0.444} & 0.500 & 0.486 \\ \bottomrule[1.5pt]
\end{tabular}}
  \vspace{-10pt}
\end{table*}

\section{Experiment Settings}
\subsection{Settings of MLLM-as-a-Judge}
We evaluate the judging performance of eleven leading MLLMs – GPT-4V~\citep{openai2023gpt4v}, Gemini-Pro-Vision-1.0 \citep{geminiteam2023gemini}, LLaVA-1.5-13b, LLaVA-1.6-7b/13b/34b \citep{liu2023llava},  Qwen-VL-Plus/Max \citep{Qwen-VL} and \text{CogVLM} \citep{wang2023cogvlm} – across three distinct evaluation settings. Adapting the ``Analyze-then-Judge'' paradigm from \citet{chiang2023closer}, which is a one-step CoT approach \citep{wei2022chain}, we first ask MLLMs to analyze responses and then provide a judgment based on their analysis. However, due to capability limitations to perform the ``Analyze-then-Judge'' setting for \text{LLaVA} and \text{CogVLM}, we prompt them to directly output their judgment. We also evaluate whether multi-step CoT will enhance the performance of MLLM serving as a judge.

Furthermore, to extensively explore MLLMs judging capabilities, we conduct experiments on various settings, including scenarios without vision input, replacing vision input with a detailed description generated by \text{GPT-4V} as a vision expert, and employing multi-step CoT. 
Considering that the first two settings do not involve image inputs, we also include tests on the latest \text{GPT-4} \citep{openai2023gpt4} \text{Gemini} \citep{geminiteam2023gemini}, \text{LLaMA-2-70b} \citep{touvron2023llama}, and \text{Mixtral-8x7b} \citep{jiang2024mixtral} to assess whether LLMs can effectively perform judging tasks without vision perception. Comprehensive details of these experimental setups are available in Appendix \ref{Detailed Experiment Settings}, and the prompts can be found in Appendix \ref{Prompt templates}.

\subsection{Judging Metrics}
After collecting responses from MLLM judgments, we quantify their alignment with human annotations across three settings, employing distinct metrics as follows:

    $\triangleright$ \textbf{Scoring Evaluation:} Following \text{LLM-as-a-Judge} \citep{zheng2023judging}, we compute the Pearson similarity \citep{lee1988thirteen} between the MLLMs' judgments and human ratings across different sub-datasets.
    
    $\triangleright$ \textbf{Pair Comparison:} We assess the similarity between the MLLM judgments and human decisions using accuracy, F1-score \citep{goutte2005probabilistic}, and recall \citep{goutte2005probabilistic} to assess the judging abilities of models.
    
    $\triangleright$ \textbf{Batch Evaluation:} We consolidate the ranking results into a singular sequence and employ the Normalized Levenshtein distance \citep{levenshtein1966binary} to evaluate the similarity between judgments from MLLMs and human annotation.

 \subsection{Human Agreement in MLLM Judgment}
Apart from traditional metrics for similarity assessment between judgments from MLLMs and humans, we further evaluate the judgments provided by MLLMs to uncover latent bias and hallucination in 10 datasets. We also invite human annotators for further validation, focusing on the following aspects:

    $\triangleright$ \textbf{Human Agreement:} This involves a simple `yes' or `no' response to assess agreement with the MLLM judgments. While some judgments might appear reasonable, they may still be considered incorrect due to unique human perspectives. Hence, we conduct experiments on human agreement to address situations that traditional metrics may not adequately capture.
    
    $\triangleright$ \textbf{Analysis Grading:} Each MLLM analysis is assigned a score from 1 to 5, considering relevance, accuracy, creativity, and response granularity, detailed in Appendix~\ref{Prompt templates}.
    
    $\triangleright$ \textbf{Hallucination Detection:} Given the propensity for hallucination issues in the complex reasoning chains and long-term vision-language contexts of MLLMs, we task human annotators with identifying any hallucinations in the analyses of MLLM judgments, adhering to established definitions of vision and language hallucination \citep{sun2024trustllm}.
    
\section{Empirical Results and Analysis}

\begin{table*}[ht]
\centering

\caption{Human agreement percentage on MLLM-as-a-Judge in 10 datasets. Each judgment is independently reviewed three times by different annotators and consensus results are recorded. \text{Gemini} failed in diffusion tasks and its results are omitted.}
\setlength{\tabcolsep}{6pt} 
\renewcommand\arraystretch{1.2}
\label{tab: human agreement}
\resizebox{0.98\linewidth}{!}{
\begin{tabular}{l l|c c c c c c c c c c c}  
\toprule[1.5pt]
\textbf{Settings} & \textbf{MLLM}  & COCO  & C.C. & Diffusion & Graphics & Math & Text & WIT & Chart & VisIT & CC-3M & Average\\
\midrule
\multirow{2}{*}{\textbf{{Score ($\uparrow$)}}}
                       & \text{Gemini} & 0.783 & \textbf{0.739} & - & 0.618 & 0.536 & 0.621 & \textbf{0.749} & 0.630 & 0.712 & 0.702 & 0.677\\
                        & \text{GPT-4V}& \textbf{0.799} & 0.725 & \textbf{0.506} & \textbf{0.688} & \textbf{0.638} & \textbf{0.706} & 0.714 & \textbf{0.676} & \textbf{0.779} & \textbf{0.754} & \textbf{0.699}\\ \midrule
\multirow{2}{*}{\textbf{{Pair ($\uparrow$)}}} 
                      & \text{Gemini} &0.705 & 0.833 & - & 0.733 & 0.520 & 0.717 & \textbf{0.827} & 0.620 & \textbf{0.853} & 0.703 & 0.724\\
                      & \text{GPT-4V} & \textbf{0.821} & \textbf{0.926} & \textbf{0.873} & \textbf{0.794} & \textbf{0.618} & \textbf{0.752} & 0.790 & \textbf{0.796} & 0.797 & \textbf{0.766} & \textbf{0.793}\\
\midrule
\multirow{2}{*}{\textbf{{Batch ($\downarrow$)}}} 
                       & \text{Gemini} & 0.642 & \textbf{0.639} & - & 0.333 & 0.330 & 0.473 & 0.511 & 0.315 & 0.422 & \textbf{0.554} & 0.469 \\
                       & \text{GPT-4V} & \textbf{0.663} & \textbf{0.639} & \textbf{0.912} & \textbf{0.536} & \textbf{0.475} & \textbf{0.615} & \textbf{0.641} & \textbf{0.640} & \textbf{0.622} & 0.467 & \textbf{0.621} \\
\bottomrule[1.5pt]
\end{tabular}}
\vspace{-10pt}
\end{table*}

\begin{figure*}[h]
    \centering
    \vspace{-2pt}
    \includegraphics[width=\linewidth]{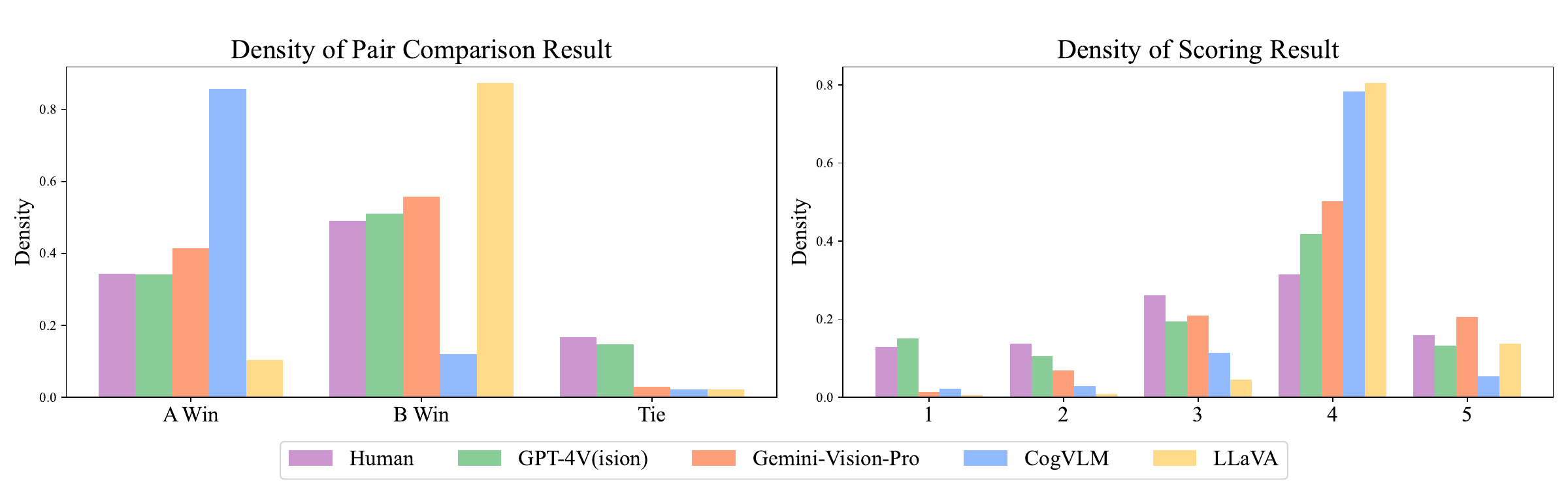}
     \vspace{-17pt}
    \caption{\textit{Pair Comparison} density (Left) and \textit{Scoring Evaluation} density (Right) of different MLLMs judgments and human annotations.}
    \label{fig: score and pair result}
     \vspace{-15pt}
\end{figure*}

\subsection{MLLM Judgment \textit{vs} Human Annotation}
As shown in Figure~\ref{fig: Radar Figure} and Table \ref{tab: human agreement}, judgments made by \text{GPT-4V} are closer to human annotations among all settings, while \text{Gemini} is far different, with \text{LLaVA}, CogVLM and Qwen-VL-Max are even worse. Overall, MLLM judgments perform better on \textit{Pair Comparison}, while falling short in \textit{Scoring Evaluation} and \textit{Batch Ranking}, showing a huge gap between the model and human preferences. Under the ``Analyze-then-Judge'' setting, \text{GPT-4V} prefers to give a longer judge in all settings, convincing its ability to reason on long-term text.

    $\triangleright$ \textbf{Scoring Evaluation:} \text{GPT-4V} demonstrates the highest similarity to human scoring with a similarity score of 0.490. In contrast, \text{Gemini} achieves only 0.304, with \text{LLaVA} and \text{CogVLM} scoring even lower. This discrepancy is mainly due to \text{Gemini}'s tendency to assign scores around 4 points as depicted in Figure \ref{fig: score and pair result}, seldom giving 1 or 2 points. \text{LLaVA} and \text{CogVLM} show a pattern similar to \text{Gemini}, predominantly assigning scores around 4 points. We attribute this to a `High-Score' Bias, akin to the `Yes/No' bias identified by \citet{liu2023hallusionbench}, which may result from an imbalance in positive and negative judging instructions in their training data \citep{liu2023aligning}, severely limits their ability to provide just and varied scores in scoring settings. In comparison, \text{GPT-4V}'s scores are more evenly distributed and align closely with human preferences.  
    
    $\triangleright$ \textbf{Pair Comparison:} As illustrated in Figure~\ref{fig: score and pair result}, \text{GPT-4V} outshines other MLLMs in pair comparison tasks, achieving 0.636 in tie settings and 0.773 in non-tie settings, surpassing 0.8 in many datasets, which indicate a strong alignment with human preferences. \text{Gemini}, \text{LLaVA}, and \text{CogVLM} show a marked preference for declaring a clear winner, possibly due to a lack of tie situations in their training, leading to biased judgments. It's also interesting that the frequency of ties given by \text{GPT-4V} closely mirrors that of human judges, suggesting similar thresholds for tie decisions.
    
    $\triangleright$ \textbf{Batch Ranking:} \text{GPT-4V} aligns more closely with human ranking results, indicating a significant lead with a mean Levenshtein Distance of 0.361. However, there is still substantial room for improvement in this task for all MLLMs. Notably, \text{CogVLM} is unable to provide a full ranking in this context, offering only the top choice; so it was excluded from this comparison; \text{LLaVA} also exhibits position bias influenced by prompt structure, often replicating judgments seen in example prompts, which complicates its ability to produce fair judgments.

\begin{table}[t]
\centering
\large
\caption{Consistency comparisons of \text{GPT-4V} and \text{Gemini} in 10 datasets. Average means weighted average for consistency times, ``MCC'' stands for ``Majority Consistency Criterion'', which deems responses consistent if over half of them are identical across our 6 repetitions of experiments.}
\renewcommand\arraystretch{1.2}
\resizebox{1\linewidth}{!}{
\begin{tabular}{l|c c|c c|c c}
\toprule[1.5pt]
\multirow{2}{*}{MLLM}& \multicolumn{2}{c|}{\textbf{Score}}  & \multicolumn{2}{c|}{\textbf{Pair}} & \multicolumn{2}{c}{\textbf{Batch}} \\
                           & Average & MCC & Average & MCC & Average & MCC \\
\midrule
\text{Gemini}            & 0.531 & 0.054 & 0.781 & 0.547 & 0.629 & 0.338 \\
\text{GPT-4V}           & \textbf{0.796} & \textbf{0.611} & \textbf{0.836} & \textbf{0.675} & \textbf{0.679} & \textbf{0.418}  \\
\bottomrule[1.5pt]
\end{tabular}}
  \vspace{-15pt}
\label{tab: consistency}

\end{table}

\subsection{MLLM Judging Consistency}
\begin{figure}
    \centering
    \includegraphics[width=1\linewidth]{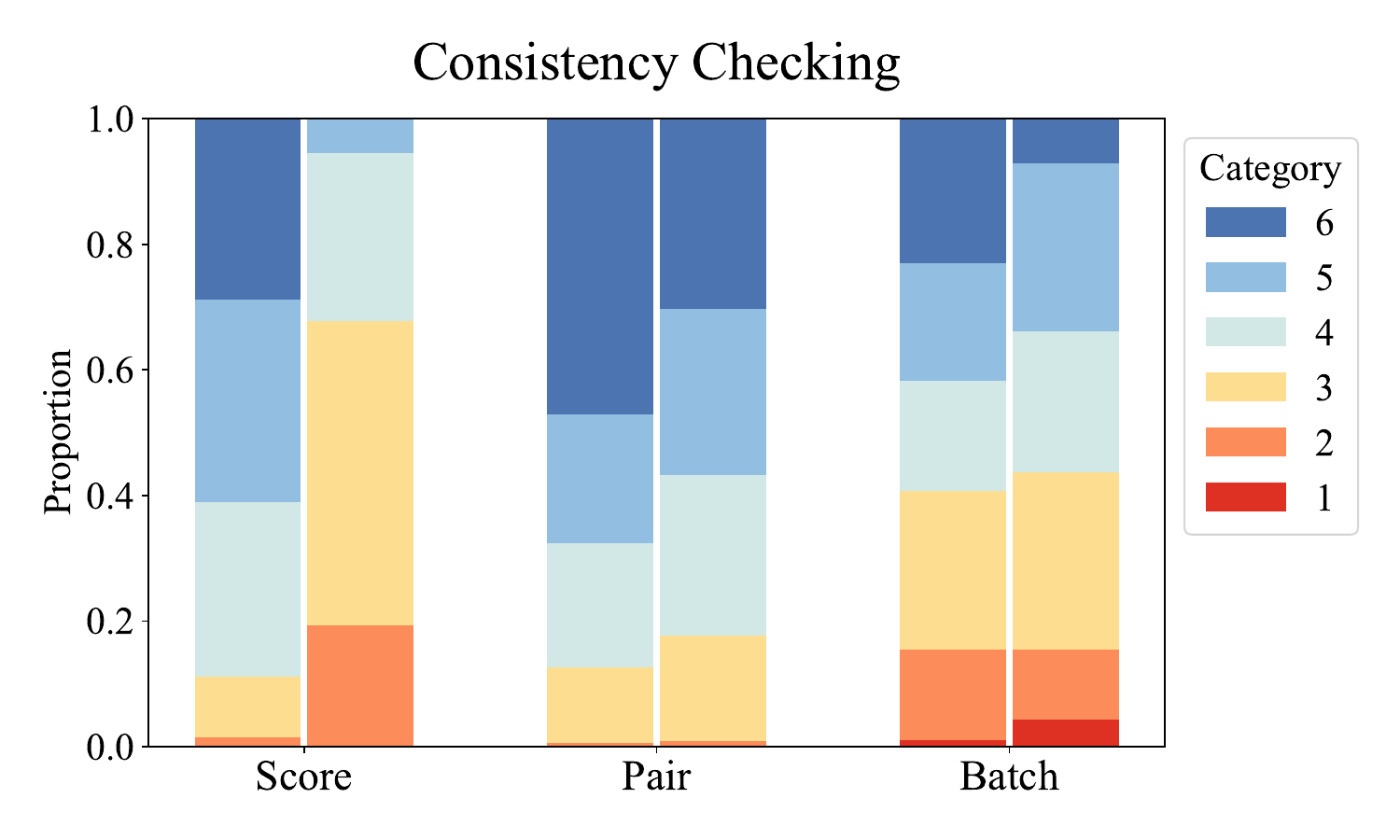}
    \vspace{-20pt}
    \caption{
    Consistency checking on 6 repetitions of experiments on \text{GPT-4V} (Left) and \text{Gemini} (Right). \text{GPT-4V} outperforms \text{Gemini} with a relatively higher ratio for high consistency.
    }
     \vspace{-15pt}
    \label{fig: consitency_bar}
\end{figure}

To be a reliable judge, consistent decision-making across repeated evaluations of the same query is crucial. For this purpose, we conduct six repeated tests with MLLM judgments and calculated the weighted average consistency scores and Majority Consistency Criterion ratios for \text{GPT-4V} and \text{Gemini}, as shown in Table~\ref{tab: consistency} and Figure~\ref{fig: consitency_bar}. Despite a higher temperature setting, \text{GPT-4V} substantially outperforms \text{Gemini} across all tasks. Particularly in \textit{Pair Comparison}, \text{GPT-4V} achieves a higher consistency score of 0.675, but it encounters difficulties in maintaining similar levels of consistency in \textit{Scoring} and \textit{Batch Ranking} tasks, with scores dropping to 0.611 and 0.418, indicating the challenge of producing qualified and convincing judgments. 

\begin{table*}[ht]
\centering

\large
\renewcommand\arraystretch{1.05}
\centering
\caption{Results of GPT-4V and Gemini-Pro acting as a judge with a 3-step CoT approach in a selected subset. 
}
\resizebox{0.98\linewidth}{!}{
\begin{tabular}{ll|ccccccccccc}\toprule[1.5pt]
\textbf{Settings} & \textbf{MLLM} & COCO & C.C. & Diffusion & Graphics & Math & Text & WIT & Chart & VisIT & CC-3M & Ave. \\ \midrule
\multirow{4}{*}{\textbf{Score ($\uparrow$)}} & GPT-4V & \textbf{0.454} & \textbf{0.507} & \textbf{0.458} & \textbf{0.645} & \textbf{0.606} & \textbf{0.624} & \textbf{0.579} & \textbf{0.645} & \textbf{0.620} & \textbf{0.431} & \textbf{0.557} \\
 & GPT-4V (+CoT) & 0.246 & 0.165 & 0.192 & 0.385 & 0.397 & 0.400 & 0.298 & 0.443 & 0.423 & 0.038 & 0.299 \\
 & Gemini & 0.262 & 0.408 & - & 0.400 & 0.228 & 0.222 & 0.418 & 0.343 & 0.336 & 0.374 & 0.299 \\
 & Gemini (+CoT) & 0.127 & 0.068 & 0.117 & 0.220 & 0.132 & 0.182 & 0.105 & 0.140 & 0.222 & 0.128 & 0.144 \\ \midrule
\multirow{4}{*}{\textbf{Pair w. Tie ($\uparrow$)}} & GPT-4V & \textbf{0.696} & \textbf{0.824} & \textbf{0.847} & \textbf{0.639} & \textbf{0.564} & \textbf{0.673} & \textbf{0.679} & \textbf{0.657} & 0.640 & \textbf{0.612} & \textbf{0.683} \\
 & GPT-4V (+CoT) & 0.507 & 0.657 & 0.561 & 0.601 & 0.515 & 0.580 & 0.489 & 0.521 & \textbf{0.646} & 0.553 & 0.563 \\
 & Gemini & 0.616 & 0.787 & - & 0.650 & 0.436 & 0.664 & 0.605 & 0.500 & 0.660 & 0.560 & 0.609 \\
 & Gemini (+CoT) & 0.233 & 0.239 & 0.420 & 0.207 & 0.284 & 0.329 & 0.352 & 0.357 & 0.247 & 0.239 & 0.291 \\ \midrule
\multirow{4}{*}{\textbf{Pair w.o. Tie ($\uparrow$)}} & GPT-4V & \textbf{0.804} & \textbf{0.870} & \textbf{0.922} & \textbf{0.807} & \textbf{0.801} & \textbf{0.805} & \textbf{0.734} & \textbf{0.849} & \textbf{0.761} & \textbf{0.703} & \textbf{0.806} \\
 & GPT-4V (+CoT) & 0.673 & 0.821 & 0.845 & 0.707 & 0.738 & 0.787 & 0.548 & 0.756 & 0.753 & 0.654 & 0.728 \\
 & Gemini & 0.717 & 0.840 & - & 0.770 & 0.678 & 0.793 & 0.688 & 0.658 & 0.711 & 0.652 & 0.723 \\
 & Gemini (+CoT) & 0.267 & 0.275 & 0.573 & 0.264 & 0.414 & 0.424 & 0.427 & 0.511 & 0.299 & 0.319 & 0.377 \\ \midrule
\multirow{4}{*}{\textbf{Batch ($\downarrow$)}} & GPT-4V & 0.323 & 0.344 & \textbf{0.092} & \textbf{0.401} & \textbf{0.367} & \textbf{0.341} & \textbf{0.302} & \textbf{0.364} & \textbf{0.313} & 0.407 & \textbf{0.325} \\
 & GPT-4V (+CoT) & 0.428 & 0.416 & - & 0.427 & 0.434 & 0.401 & 0.366 & 0.406 & 0.422 & 0.472 & 0.419 \\
 & Gemini & \textbf{0.287} & \textbf{0.299} & - & 0.473 & 0.462 & 0.430 & 0.344 & 0.520 & 0.426 & \textbf{0.357} & 0.400 \\
 & Gemini (+CoT) & 0.441 & 0.481 & 0.542 & 0.595 & 0.494 & 0.533 & 0.483 & 0.569 & 0.486 & 0.463 & 0.509 \\
 \bottomrule[1.5pt]
\end{tabular}}
  \vspace{-10pt}
  \label{tab: COT result}
\end{table*}

\subsection{Human Agreement}
Our manual evaluation of MLLMs on agreement and scoring, revealed notable findings. Table \ref{tab: human agreement} shows that GPT-4V achieved around 70\% human agreement across all settings, excelling in the \textit{Pair Comparison} task with 79.3\% agreement. Specifically, GPT-4V reached 78\% in human agreement for \textit{Pair Comparison}, with Gemini close at 72\%, indicating strong performance in most sample pairs and supporting the idea that large models excel in pairwise distinctions \citep{zheng2023judging}, though improvements are needed in other judging settings.

In \textit{Scoring Evaluation}, GPT-4V achieves a 70\% human agreement rate, peaking at 79.9\% in MS-COCO, while Gemini averaged 67.7\%. To assess the consistency of MLLM judging quality across multiple responses to a single image-instruction pair, we use Mean Absolute Deviation (MAD) metric to measure the average absolute variance between individual scores and the mean. 
Figure~\ref{fig:MAD} shows that GPT-4V exhibits lower variation in quality assessments, indicating more consistent and reliable judgment compared to Gemini. However, in \textit{Batch Ranking}, both models exhibited decreased alignment with human judgments, especially in Maths and graphic information processing, suggesting that models may lack the capabilities to fully comprehend user instructions, leading to less reliable judgments.

\subsection{Multi-steps CoT Do Not Enhance Performance}
We have conducted additional tests using GPT-4V and Gemini with a 3-step CoT approach for judging, as detailed in Table~\ref{tab: COT result}. Our analysis reveals that while employing CoT with additional steps markedly reduces hallucinations in judgments, it does not align more closely with human preferences. On numerous datasets, this approach even diminishes judging performance. Specifically, Gemini's effectiveness drops more drastically. With 3-step CoT, there is an increased likelihood that the judgment will be disturbed by its understanding of the figure and its own responses to the instruction, thereby undermining its final judgment if hallucinations exist in the previous chain.

\subsection{Vision Perception Benefits MLLM Judging}
We explore the feasibility of using LLMs for judging text-based responses without directly analyzing the original images. This involves two approaches: omitting vision information entirely and providing a detailed description of the picture. We choose LLaMA-70b, Mixtral8x7b-v0.1 and GPT-3.5 to provide descriptions. 
Surprisingly, as illustrated in Table~\ref{tab: ablation study on llm}, we find that LLMs' performance in multimodal judging tasks significantly improve with picture descriptions, achieving a Pearson similarity of 0.435 in \textit{Scoring Evaluation} tasks, markedly outperformed judgments made without any vision perception. Notably, in no-tie \textit{Pair Comparison}, MLLMs with detailed vision descriptions even exceed the standard performance of MLLMs in judging. This suggests that MLLMs may lack certain human-like judging capabilities, while LLMs can be potential judges for multimodal tasks when provided with comprehensive task-related descriptions.

\begin{table}[t]
\vspace{-0.5em}
\renewcommand\arraystretch{1.1}
\caption{How vision perception significantly enhances multimodal judging performance in traditional LLM-as-a-Judge setting, slightly outperforming MLLMs in judging. Vision Exp. stands for judging with a detailed image description. }
\label{tab: ablation study on llm}
\resizebox{1\linewidth}{!}{
\begin{tabular}{llc|cc|c}
\toprule[1.5pt]
\multirow{2}{*}{\textbf{MLLM}} & \multirow{2}{*}{\textbf{Settings}} & \textbf{Score ($\uparrow$)} & \multicolumn{2}{c|}{\textbf{Pair ($\uparrow$)}} & \textbf{Batch ($\downarrow$)} \\
 &  & Pearson & w. Tie & w.o. Tie & Edit Dis. \\ \midrule
\multirow{2}{*}{\textbf{LLaMA2-70b}} & Vision Exp & 0.060 & 0.404 & 0.550 & 0.643 \\
 & No Vision & 0.126 & 0.374 & 0.537 & 0.583 \\ \midrule
\multirow{2}{*}{\textbf{Mixtral-8x7b}} & Vision Exp & 0.054 & 0.374 & 0.543 & 0.603 \\
 & No Vision & 0.151 & 0.478 & 0.731 & 0.546 \\ \midrule
\multirow{2}{*}{\textbf{GPT-3.5}} & Vision Exp & 0.154 & 0.453 & 0.591 & 0.473 \\
 & No Vision & 0.223 & 0.459 & 0.644 & 0.504 \\ \midrule
\multirow{2}{*}{\textbf{GPT-4V}} & Vision Exp & \textbf{0.435} & \textbf{0.544} & \textbf{0.878} & 0.400 \\
 & No Vision & 0.299 & 0.491 & 0.868 & \textbf{0.394} \\ \midrule
\multirow{2}{*}{\textbf{Gemini}} & Vision Exp & 0.120 & 0.438 & 0.785 & 0.472 \\
 & No Vision & 0.108 & 0.433 & 0.758 & 0.470 \\ \bottomrule[2pt]
\end{tabular}}
  \vspace{-20pt}
\end{table}

\subsection{Bias and Hallucination}
\begin{figure*}[h]
    \centering
    \vspace{-3pt}
    \includegraphics[width=\linewidth]{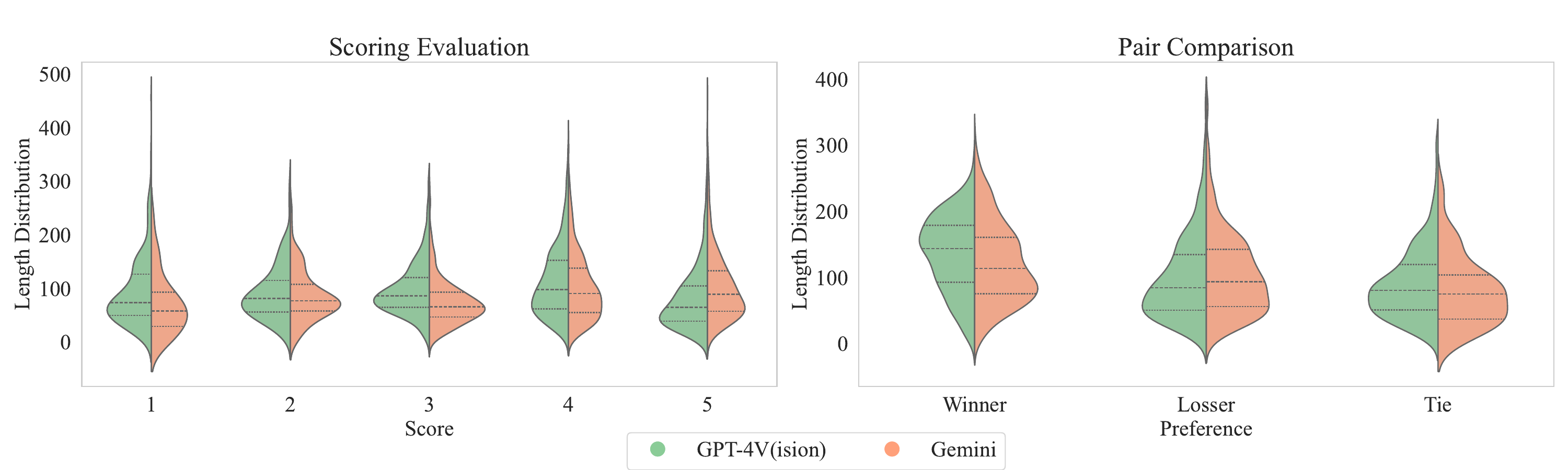}
    \vspace{-10pt}
    \caption{Length bias in 10 datasets. The horizontal axis represents length, and the vertical axis represents density.}
    \vspace{-10pt}
    \label{fig: length_bias}
\end{figure*}
\begin{figure*}[h]
    \centering
    \includegraphics[width=\linewidth]{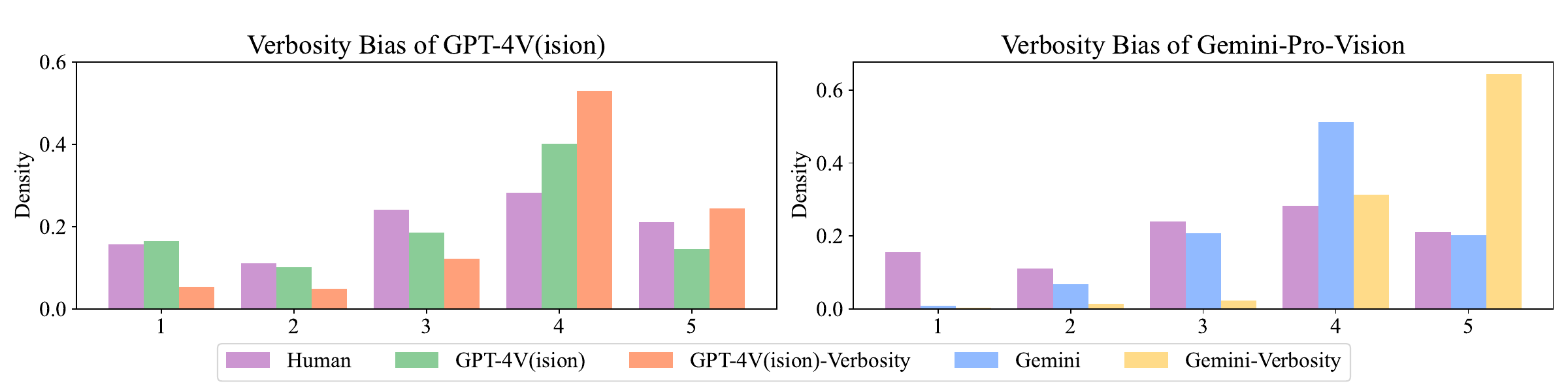}
      \vspace{-15pt}
    \caption{Length Bias in Different MLLM judgments.}
      \vspace{-10pt}
    \label{fig: Verbosity Bias}
\end{figure*}
    \paragraph{Egocentric Bias.}
    Models tend to assign higher scores to their own responses while scoring others lower \citep{zheng2023judging, li2024leveraging}. In Figures~\ref{fig: Ego Bias} and \ref{fig: Ego Bias on pair}, \text{GPT-4V} exhibits a slight degree of Egocentricity. Conversely, Gemini maintains a uniform scoring distribution across different sources, demonstrating a more equitable approach to judgment. In contrast, GPT-4V shows self-preference, aligning its judgments with its predefined ethical guidelines. For example, GPT-4V consistently emphasizes privacy preservation, leading to higher scores for privacy-related questions based on its own metrics. Despite efforts in prompt engineering to ensure neutrality, these models still rely on judgment criteria set during post-alignment training \citep{ouyang2022training}. This bias can result in judgments that deviate from human preferences, highlighting the complexity of aligning MLLM judgments with humans'.
    
    \paragraph{Position Bias.}
    Model consistently favor answers in specific positions, often influenced by training data that typically places correct responses at the beginning or end of prompts \citep{liu2023lost}. Figure \ref{fig: score and pair result} illustrates bias in LLaVA and CogVLM during Pair Comparison tasks, where they consistently prefer answers in a specific position. This bias likely arises from their limited ability to follow complex instructions, leading them to be influenced by prompt structure. For example, if a \textit{Batch Ranking} prompt includes a sequence like `ABCD’, LLaVA replicates this sequence in 88.2\% of responses, significantly more than other sequences. However, this bias can be reduced by introducing multiple examples, suggesting that prompts with more examples can better direct these models to follow instructions accurately.
    
    \paragraph{Length Bias.}
    Models tend to prefer longer answers over concise but correct ones \citep{li2024leveraging}, also known as verbosity bias \citep{zheng2023judging}. Figure~\ref{fig: length_bias} shows that both GPT-4V and Gemini assign higher scores to longer content. We conducted an expanded scoring experiment using GPT-4 \citep{openai2023gpt4} without vision, increasing the semantic length of answers without changing their original intent. In Figure~\ref{fig: Verbosity Bias}, we observe noticeable score increases, with GPT-4V and Gemini showing average gains of 0.6 and 0.75 points, respectively. These results suggest that MLLMs may favor longer text for higher scores.

\noindent\textbf{Hallucination Detection and Mitigation.}
We observe a higher frequency of hallucinations in \textit{Batch Ranking}, compared to \textit{Pair Comparison} and \textit{Scoring Evaluation}. These hallucinations involved significant misinterpretations and retrieval errors, impacting judgment accuracy and reliability. To address this, we employed a multi-step CoT approach on \textsc{MLLM-as-a-Judge-Hard}, adding reasoning steps before the conventional ``Analyze-then-Judge'' process. This enhanced procedure included: 1) image-instruction, 2) image, and 3) instruction. In Table~\ref{tab: COT score}, this strategy effectively reduced hallucinations across all formats, with significant improvements in tasks involving image-related information. In the \textit{Batch Ranking} task, which requires handling longer text sequences, the detailed reasoning steps were particularly effective in reducing hallucinations.

\subsection{Scaling Law for MLLM-as-a-Judge}
We conduct two sets of experiments with models of different sizes, the LLaVA-1.6 series models and the Qwen series models in four newly added datasets, illustrated in Figure \ref{fig: rebuttal figure1} and \ref{fig: rebuttal figure2}. In \textit{Score evaluation}, LLaVA-1.6-34b and Qwen-VL-Max slightly outperform others in Math, Chart, and Text tasks, showing a relatively strong scaling law. 

\begin{table}[]
\vspace{-0.5em}
\centering
\renewcommand\arraystretch{1.1}
\caption{
Reduction of hallucinations in \textsc{MLLM-as-a-Judge-Hard} through additional CoT steps compared to normal setting.
}
\label{tab: COT score}
\resizebox{0.9\linewidth}{!}{
\begin{tabular}{cccc}
\toprule[1.5pt]
\textbf{Setting} & \begin{tabular}[c]{@{}c@{}}Figure-\\ instruction\end{tabular} & Figure & Instruction \\ \midrule
\textbf{Score} & 46.15\%                                                                 & \textbf{48.72\%}  & 33.33\%        \\
\textbf{Pair}  & 28.21\%                                                                 & \textbf{35.90\%}   & 33.33\%        \\
\textbf{Batch} & \textbf{43.59\%}                                                                 & 35.90\%   & 35.90\%        \\ \bottomrule[1.5pt]
\end{tabular}}
\vspace{-16pt}
\end{table}

\section{Related Work}
\paragraph{LLM as a Judge.}
The evolution of LLMs has made them increasingly effective evaluators in Natural Language Processing (NLP) tasks. \citet{zhu2023judgelm} introduced JudgeLM for LLM evaluation, followed by AUTO-J \citep{li2023generative}, aligning closely with human judgment \citep{bai2023touchstone, li2023alpacaeval, kim2023prometheus}. Advancements in CoT reasoning \citep{wei2022chain, chu2023survey} and training-free instruction following \citep{brown2020language, wei2021finetuned} further extend LLMs' judging capability in diverse tasks like translation quality assessment \citep{kocmi2023large} and story generation \citep{chiang2023can}.

\paragraph{Hallucination and Bias in Judgments.}
MLLMs suffer from vision and language hallucinations \citep{ji2023survey, huang2023survey, cui2023holistic, wang2023evaluation}, often due to vision-language misalignments in training phase~\citep{sun2024trustllm, huang2023trustgpt}. Recent research focuses on hallucination evaluation \citep{liu2023hallusionbench}, detection \citep{li2023evaluating, wang2023evaluation}, and mitigation \citep{yin2023woodpecker, gunjal2023detecting, zhou2023analyzing}, noting that even GPT-4V suffer from these issues \citep{shi2023exploring, liu2023hallusionbench, cui2023holistic}. Besides, biases in MLLM-as-a-Judge, similar to those in human decision-making \citep{blunch1984position, raghubir2006center} and other ML domains \citep{wang2018position, liu2023lost}, such as position \citep{zheng2023large}, egocentric \citep{li2024leveraging}, and verbosity biases \citep{saito2023verbosity}, are compounded by the integration of visual perception, necessitating further investigation.

\section{Future Directions}
\paragraph{Multimodal RLHF/DPO.}
Our work is highly connected with multimodal RLHF/DPO \citep{sun2023aligning, li2023silkie, yu2023rlhf}. Our dataset includes extensive human annotations, such as manually assigned scores and preference on pairs, which could serve as invaluable training material for RLHF reward models and supply paired data essential for DPO \citep{rafailov2024direct, zhang2024direct}, paving the way for enhancing the training of MLLMs.

\paragraph{Exploring the upper bound of MLLM-as-a-Judge.}
Beyond expanding the steps in the Chain of Thought prompting \citep{wei2022chain}, we see significant potential in more sophisticated reasoning frameworks, such as multi-agent debating \citep{chan2023chateval} when MLLM acts as a Judge, which could enhance the judging accuracy through improved reasoning capabilities. Additionally, addressing inherent biases in the model during the judgment process is crucial. For instance, position bias in \textit{Pair Comparison} and \textit{Batch Ranking} \citep{zheng2023large, wang2024my}, and the tendency to assign higher scores, as discussed in \citep{lee2024prometheus}, are critical areas for improvement.

Incorporating a human-in-the-loop approach \citep{wang2023large} offers a promising solution to enhance judgment consistency and reliability. For example, if judgment results vary in more than half of several repeated judgments, it may need human intervention for consistency checking. When it's challenging to discern the MLLM's judgment due to non-compliance with the suggested output format or lack of a clear outcome, human intervention may be required to refine this process by manually verifying judgments. 

\section{Conclusion}
In this paper, we have presented a new benchmark, termed {MLLM-as-a-Judge}, to assess the judging capabilities of MLLMs across three critical evaluation settings in the multimodal domain: \textit{Scoring Evaluation}, \textit{Pair Comparison}, and \textit{Batch Ranking}. We further evaluate their agreement with humans. Our results reveal that advanced MLLMs can win significant human recognition in \textit{Pair Comparisons}, but perform poorly in \textit{Scoring Evaluation} and \textit{Batch Ranking} tasks. Our work highlights potential areas for future refinement and improvement of MLLMs.
We advocate for additional efforts dedicated to supporting the continuous development of MLLMs as judges.



\section*{Impact Statement}
In this paper, we introduce a novel benchmark, termed \text{MLLM-as-a-Judge}, designed to propel the evolution of MLLMs toward achieving judgments that align more closely with human perspectives. This benchmark establishes a heightened criterion for assessing MLLMs, emphasizing their proficiency in comprehending and processing information in a manner reflective of human cognitive processes.  One limitation of our work lies in the bias in human annotation and MLLMs. We leave the exploration of more objectives, ethically principled, and socially beneficial MLLM-as-a-Judge systems as our future work.

\bibliography{custom}
\bibliographystyle{icml2024}

\newpage
\appendix
\onecolumn
\section{Comprehensive Related Works}
\label{Full Related Works}
\subsection{Large Model as Judge}
The rapid development of LLMs has significantly enhanced their capabilities in long-term context perception and reasoning, increasingly popularizing their use as evaluators in various Natural Language Processing (NLP) tasks. \citet{zhu2023judgelm} were pioneers in this area, introducing JudgeLM, a fine-tuned LLM designed for evaluating other LLMs. Building on this, \citet{li2023generative} introduced AUTO-J, a system that evaluates LLMs through both pairwise comparisons and single-response assessments, demonstrating close alignment with human judgment \citep{bai2023touchstone, li2023alpacaeval, kim2023prometheus}. Further advancements in LLMs, such as the development of Chain-of-Thought reasoning \citep{wei2022chain, chu2023survey}, training-free instruction following \citep{brown2020language, wei2021finetuned}, and enhanced alignment with human preferences \citep{ouyang2022training}, have solidified their role in diverse tasks like translation quality assessment \citep{kocmi2023large} and story generation \citep{chiang2023can}.

\subsection{Hallucination and Bias in Judge}
MLLMs are known to exhibit both vision hallucination and hallucination originating from LLMs, a phenomenon typically characterized by responses containing information not present in the visual or natural language context \citep{ji2023survey, huang2023survey, cui2023holistic, wang2023evaluation}. This issue often stems from misalignments in vision-language training \citep{sun2024trustllm, huang2023trustgpt}. Recent studies have begun to address these hallucination issues, focusing on evaluation \citep{liu2023hallusionbench}, detection \citep{li2023evaluating, wang2023evaluation}, and mitigation strategies \citep{yin2023woodpecker, gunjal2023detecting, zhou2023analyzing}. Notably, GPT-4V \citep{openai2023gpt4v}, despite being a leading model in many fields \citep{yang2023dawn, wu2023early}, has also demonstrated susceptibility to hallucinations \citep{shi2023exploring, liu2023hallusionbench, cui2023holistic}. This raises concerns about the reliability of MLLMs in evaluative roles.

In terms of bias, MLLM judging is subject to issues not exclusive to our context of evaluation but also observed in human decision-making \citep{blunch1984position, raghubir2006center} and Machine Learning (ML) domains \citep{wang2018position, liu2023lost, huang2024decision} such as position bias \citep{zheng2023large}, egocentric bias \citep{li2024leveraging}, and verbosity bias \citep{saito2023verbosity}. The integration of visual perception in MLLMs introduces additional complexities, resulting in biases unique to the fusion of dual perceptions, an area that still demands thorough exploration.

\subsection{Evaluating Large Multimodal Models}
Evaluating MLLMs typically involves diverse tasks and corresponding metrics, which reflect the models' ability to comprehend and generate content based on both visual and textual information. For instance, in image captioning tasks, models are tasked with generating descriptive text for a given image. The effectiveness of these models is measured using metrics such as BLEU \citep{papineni2002bleu}, METEOR \citep{banerjee2005meteor}, ROUGE \citep{lin2004rouge}, and CIDEr \citep{vedantam2015cider}. In the context of Visual Question Answering (VQA), models are evaluated based on their ability to answer questions on an image’s content. Here, the accuracy of model responses is compared against human-annotated answers, serving as the primary metric \citep{antol2015vqa} to ensure alignment with human preferences.

However, when tackling sophisticated visual-language tasks, conventional evaluation metrics often fail to accurately capture the nuanced responses generated by these models, especially in complex or subjective scenarios that involve both visual elements and extended textual content \citep{liu2023hallusionbench}. Additionally, while manual annotation offers a more comprehensive and human-like evaluation, it comes with significant challenges. These include high costs \citep{Prendki2023are}, potential biases \citep{zheng2023judging}, and the difficulty of ensuring consistent replication \citep{chiang2023can}. These limitations highlight the need for a more holistic approach to evaluation, one that combines human-like calibration with more fine-grained assessment methods.

\section{Detailed Benchmark Construction}
\label{Detailed Benchmark Construction}

\subsection{Step 1: Image-Instruction Collection}
To attain the outlined objectives, our approach begins with a detailed analysis of the capabilities of MLLMs. Specifically, we focus on the following abilities within MLLMs:
\begin{itemize}[nolistsep, leftmargin=*]
    \item \textbf{Recognition Ability}: This encompasses general visual recognition capabilities, such as object recognition, Optical Character Recognition (OCR), and other high-level tasks in computer vision \cite{Yu2023MMVetEL}.
    \item \textbf{Comprehension Ability}: This pertains to the model's proficiency in spatial understanding and scenario comprehension.
    \item \textbf{Inferential Ability}: This involves the model's capacity to process information and reasoning, a critical component in processing charts, graphs, and mathematics.
    \item \textbf{Multilingual Ability}: This assesses the model's competence in understanding and processing multiple languages, especially focusing on their appearance in visual tasks such as text reading on images \citep{Singh2019TowardsVM}.
\end{itemize}

To ensure a robust and comprehensive assessment, we meticulously identify and incorporate 10 diverse datasets \ref{Step1: Detailed Dataset} into our evaluation framework. This strategic selection aims to enrich the diversity of our assessment tasks and enhance the breadth and depth of our evaluation capabilities, as well as prevent biases. These datasets are chosen based on their ability to effectively challenge the various aspects of MLLMs, via different downstream tasks, ensuring a thorough and nuanced understanding of their performance and potential.

To construct a robust and unbiased set of image-instruction pairs, we randomly select 300 images from each dataset, ensuring a diverse representation. Specifically, for the MathVista dataset, which includes the provision of hints, we extract 600 corresponding instructions, encompassing both scenarios: with and without hints. For the remaining datasets, we align 300 instructions with the sampled images. This process culminates in a comprehensive collection comprising 4,114 images corresponding with 4,414 instructions.

\begin{table*}[ht]
\renewcommand\arraystretch{0.7}
\centering
\caption{Datasets and corresponding tasks in benchmark construction, each task is matched with several required abilities. (Rec.-Recognition, Comp.-Comprehension, Inf.-Inferential, Mul.-Multilingual)}
\label{Step1: Detailed Dataset}
\scalebox{0.8}{
\begin{tabular}{l|c c c c l l}
\toprule[2pt]
\multirow{2}{*}{\textbf{\large Dataset}} & \textbf{\large Image} & \multirow{2}{*}{\textbf{\large \#Images}} & \multirow{2}{*}{\textbf{\large \#Questions}} & \textbf{\large \#Selected} & \multirow{2}{*}{\textbf{\large Task}} & \textbf{\large Ability} \\
& \textbf{\large Type} & & & \textbf{\large Pairs} & & \textbf{\large Required} \\
\midrule
\multirow{2}{*}{\textbf{Conceptual Captions}} & \multirow{2}{*}{Web image} & \multirow{2}{*}{3.3M} & \multirow{2}{*}{--} & \multirow{2}{*}{300} & \multirow{2}{*}{Captioning} & \multirow{2}{*}{Rec.\&Comp.} \\
& & & & & & \\
\cite{Sharma2018ConceptualCA} & & & & & & \\
\multirow{2}{*}{\textbf{ChartQA}} & \multirow{2}{*}{Chart} & \multirow{2}{*}{21K} & \multirow{2}{*}{33K} & \multirow{2}{*}{300} & \multirow{2}{*}{Chart reasoning} & \multirow{2}{*}{Rec.\&Comp.} \\
& & & & & & \\
\cite{masry-etal-2022-chartqa} & & & & & & \\
\multirow{2}{*}{\textbf{InfographicVQA}} & \multirow{2}{*}{Infographics} & \multirow{2}{*}{5.4K} & \multirow{2}{*}{30K} & \multirow{2}{*}{300} & \multirow{2}{*}{Graph reasoning} & \multirow{2}{*}{Rec.\&Comp.} \\
& & & & & & \\
\cite{Mathew2021InfographicVQA} & & & & & & \\
\multirow{2}{*}{\textbf{MathVista}} & \multirow{2}{*}{Mathematics} & \multirow{2}{*}{6K} & \multirow{2}{*}{6K} & \multirow{2}{*}{300} & \multirow{2}{*}{Math reasoning} & \multirow{2}{*}{Rec.\&Comp.\&Inf.} \\
& & & & & & \\
\cite{Lu2023MathVistaEM} & & & & & & \\
\multirow{2}{*}{\textbf{TextVQA}} & \multirow{2}{*}{Text} & \multirow{2}{*}{28K} & \multirow{2}{*}{45K} & \multirow{2}{*}{300} & \multirow{2}{*}{Text reading} & \multirow{2}{*}{Rec.\&Comp.} \\
& & & & & & \\
\cite{Singh2019TowardsVM} & & & & & & \\
\multirow{2}{*}{\textbf{WIT}} & \multirow{2}{*}{Multilingual text} & \multirow{2}{*}{11.5M} & \multirow{2}{*}{--} & \multirow{2}{*}{300} & \multirow{2}{*}{Transcription} & \multirow{2}{*}{Rec.\&Mul.} \\
& & & & & & \\
\cite{Srinivasan2021WITWI} & & & & & & \\
\multirow{2}{*}{\textbf{MS COCO}} & \multirow{2}{*}{Real-life scene} & \multirow{2}{*}{328K} & \multirow{2}{*}{2.5M(labels)} & \multirow{2}{*}{300} & \multirow{2}{*}{Image Segmentation} & \multirow{2}{*}{Rec.\&Comp.} \\
& & & & & & \\
\cite{Lin2014MicrosoftCC} & & & & & & \\
\multirow{2}{*}{\textbf{DiffusionDB}} & \multirow{2}{*}{Diffusion} & \multirow{2}{*}{14M} & \multirow{2}{*}{1.8M(prompts)} & \multirow{2}{*}{300} & \multirow{2}{*}{Comprehensive} & \multirow{2}{*}{Rec.\&Comp.\&Inf.} \\
& & & & & & \\
\cite{Wang2022DiffusionDBAL} & & & & & & \\
\multirow{2}{*}{\textbf{CC-3M Concept-balanced}} & \multirow{2}{*}{Comprehensive} & \multirow{2}{*}{595K} & \multirow{2}{*}{595K} & \multirow{2}{*}{300} & \multirow{2}{*}{Comprehensive} & \multirow{2}{*}{Rec.\&Comp.\&Inf.} \\
& & & & & & \\
\cite{liu2023llava} & & & & & & \\
\multirow{2}{*}{\textbf{VisIT-Bench}} & \multirow{2}{*}{Comprehensive} & \multirow{2}{*}{1K} & \multirow{2}{*}{592} & \multirow{2}{*}{300} & \multirow{2}{*}{Instruction following} & \multirow{2}{*}{Rec.\&Comp.\&Inf.} \\
& & & & & & \\
\cite{Bitton2023VisITBenchAB} & & & & & & \\
\multirow{2}{*}{\textbf{Mind2Web}} & \multirow{2}{*}{Webpage} & \multirow{2}{*}{2K} & \multirow{2}{*}{2K} & \multirow{2}{*}{300} & \multirow{2}{*}{Website Understanding} & \multirow{2}{*}{Rec.\&Comp.\&Inf.} \\
& & & & & & \\
\cite{deng2024mind2web} & & & & & & \\
\multirow{2}{*}{\textbf{AesBench}} & \multirow{2}{*}{Aesthetics Perception} & \multirow{2}{*}{3K} & \multirow{2}{*}{8K} & \multirow{2}{*}{300} & \multirow{2}{*}{Aesthetics Perception} & \multirow{2}{*}{Rec.\&Comp.\&Inf.} \\
& & & & & & \\
\cite{huang2024aesbench} & & & & & & \\
\multirow{2}{*}{\textbf{ScienceQA}} & \multirow{2}{*}{Science Knowledge} & \multirow{2}{*}{21K} & \multirow{2}{*}{21K} & \multirow{2}{*}{300} & \multirow{2}{*}{Reasoning} & \multirow{2}{*}{Comp.\&Inf.} \\
& & & & & & \\
\cite{lu2022learn} & & & & & & \\
\multirow{2}{*}{\textbf{MMvet}} & \multirow{2}{*}{Comprehensive} & \multirow{2}{*}{214} & \multirow{2}{*}{214} & \multirow{2}{*}{214} & \multirow{2}{*}{Instruction following} & \multirow{2}{*}{Rec.\&Comp.\&Inf.} \\
& & & & & & \\
\cite{Yu2023MMVetEL} & & & & & & \\
\bottomrule[2pt]
\end{tabular}
}
\end{table*}

\subsection{Step 2: MLLM Responses Collection}
\label{Detailed Benchmark Construction: Step 2}
\begin{figure*}[h]
    \centering
    \includegraphics[width=\linewidth]{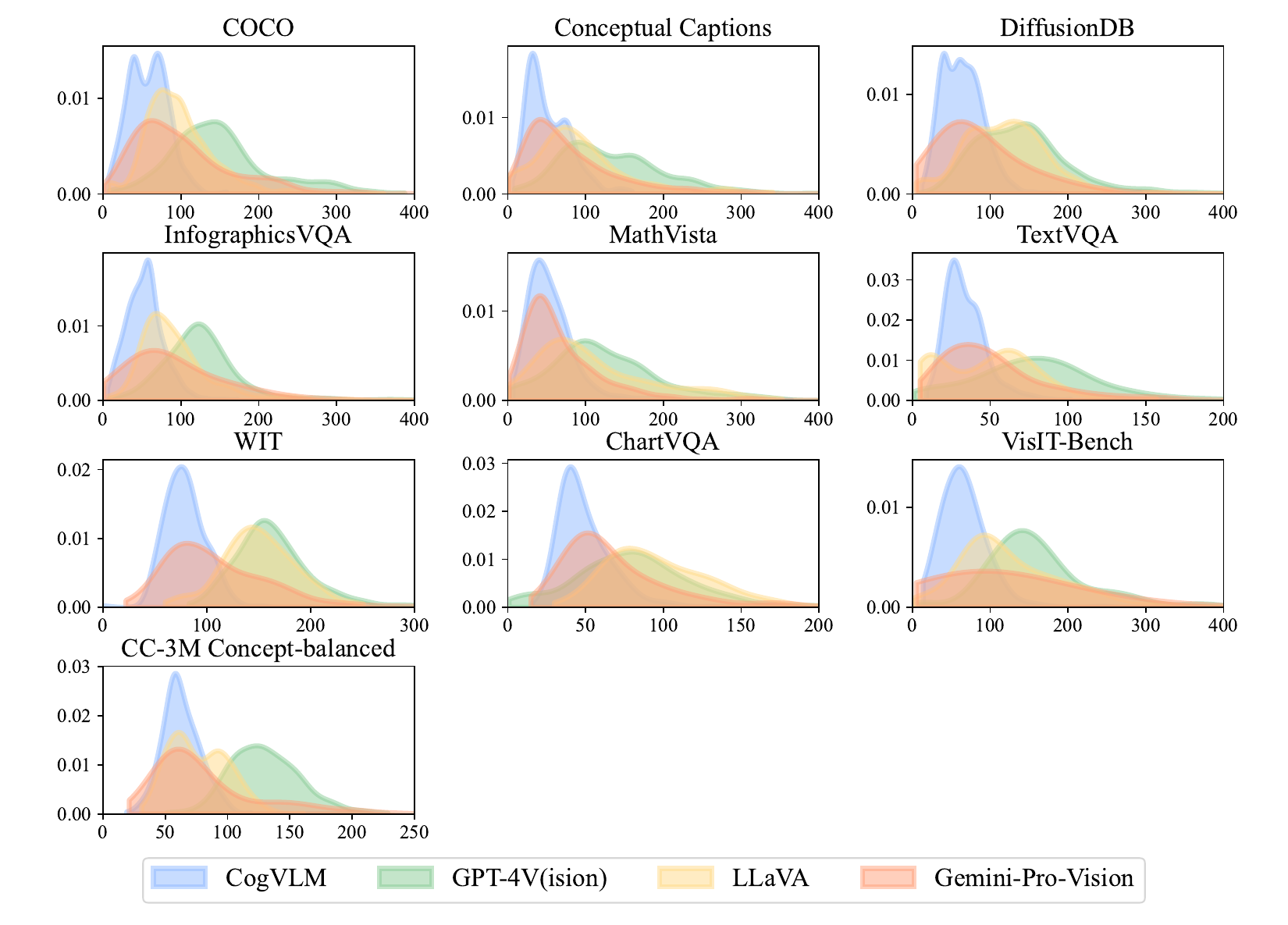}
    \vspace{-15pt}
    \caption{Response length distribution in 10 datasets. The horizontal axis represents length, and the vertical axis represents density.}
    \label{fig:length_dataset}
\end{figure*}

We engage with 4 mainstream MLLMs (i.e., GPT-4V, Gemini, LLaVA, CogVLM) by providing them with our assembled image-instruction pairs for the first 3,300 image-instruction pairs, each VLM generated a response, resulting in a comprehensive collection of 13,200 answers, with each of the 3,300 instructions receiving a distinct response from each of the four MLLMs. For the last 4 datasets, we added during the rebuttal, we leverage GPT-4V, Gemini, Qwen-VL-Max, and LLaVA-1.6-34b. For the sequential dataset Mementos \citep{wang2024mementos}, we leverage GPT-4V, Qwen-VL-Max, ChatUnivi \citep{jin2023chat}, VideoChat2 \citep{li2023mvbench} to generate responses. 
Upon collecting a total of 17,656 responses from the MLLMs, we proceed to analyze the distribution of response lengths for each model. Figure~\ref{fig:length_dataset} is a detailed illustration of length distribution in corresponding datasets.

\section{Detailed Experiment Settings}
\label{Detailed Experiment Settings}
\subsection{Response VLM Settings}
We use \text{GPT-4V}, \text{Gemini}, \text{LLaVA-1.5-13b}, \text{CogVLM}, \text{Qwen-VL-Max}, \text{LLaVA-1.6-34b} to answer the image-instruction pair. We discuss their hyperparameter settings and problems encountered during inference respectively:
\begin{itemize}
    \item \textbf{GPT-4V} \citep{openai2023gpt4v}. We set the temperature and top-p as 0.9, max-token as 2048. However, we encounter some situations where it cannot answer accurately or refuses to answer due to ethical issues like \textit{Unfortunately, due to my programming, I'm unable to ...}, which brings some difficulties to us in defining its judging capability.
    \item \textbf{Gemini} \citep{geminiteam2023gemini}. We use the default settings, which set temperature as 0.4, top-p as 1, and max-token as 2048. It should be noted that \text{Gemini} will receive more ethical limitations than GPT-4V, and will refuse to answer on the diffusion data set. But for some more difficult questions, it can't answer the questions, but it will "forcibly answer" the user's questions. In this case, GPT-4V will sincerely admit its shortcomings and give a possible answer.
    \item \textbf{LLaVA-1.5-13b} \citep{liu2023llava}. We set temperature as 0, tok-$p$ as 1, max-token as 2048, and beam search number as 3. The reason why we select such a low temperature is that \text{LLaVA} cannot correctly output its judge in a specific format. We collect responses by inference on a dual-4090 local server.
    \item \textbf{CogVLM} \citep{wang2023cogvlm}. For the hyper-parameter, we use the default setting and set max-token as 2048. We collect responses by inference on a dual-4090 local server.
    \item \textbf{Qwen-VL Family} \citep{Qwen-VL}. We use the default settings for Qwen-VL family, with top-p as 0.8 and max-token as 2048.
    \item \textbf{LLaVA-1.6 Family} \citep{liu2023improvedllava}. We set the temperature as 0.4 and top-p as 0.9, max-token as 2048. 
\end{itemize}
\subsection{GPT-4V as Judge}
We adopt GPT-4V without using JSON Mode mod based on our preliminary experiment in Appendix~\ref{Human Agreement on GPT-4V Output Mode}, but required it to output in JSON format in our prompt. Following the hyper-parameter set in \citep{chiang2023closer}, we set the temperature to 0.9, top-$k$ to 0.9, and max-token to 2048 in both cases with and without pictures. When there is a format error in the output or ethical settings are triggered, we will sample again. If it is still wrong, we will skip this piece of data.
\subsection{Gemini-Vision-Pro as Judge}
\begin{figure}
\centering

\begin{tcolorbox}[colback=gray!0!white,colframe=gray!50!black,   colbacktitle=gray!75!black]
 \begin{center}
    \includegraphics[width = 0.4\linewidth]{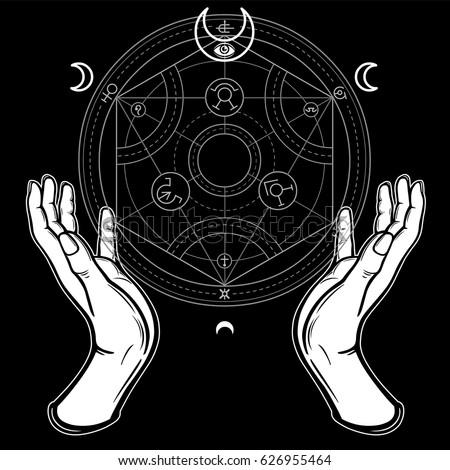}
    \end{center}

\textbf{Instruction:} Here is the caption of this figure: human hands touch an alchemical circle. According to this figure and the caption, please give a detailed analysis of this figure: \\

\textbf{Answer:} 
\begin{center}
    \includegraphics[width = 0.7\linewidth]{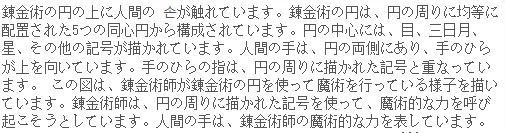}
    \end{center}

\textbf{Analysis:} The assistant's response is in \textbf{Japanese}. It provides a detailed analysis of the figure, including the symbols on the alchemical circle and the possible meaning of the hands touching it. The response is well-written and easy to understand."\\
\textbf{Judgment:} 5

\end{tcolorbox}
\caption{Gemini speaks Japanese.}
\label{gemini_chinese}
\end{figure}

We call Gemini's API on the Google Cloud Server and use the default settings for temperature and top-$k$.
It should be noted that even though Gemini is currently the only VLM that can perform Judge, it will occasionally speak Chinese or Japanese during the output process when using English text and English pictures, shown as Unicode in Figure \ref{gemini_chinese}. Although it is semantically continuous, that is, it can be used after translation, and it greatly affects the continuity of the judge. We believe that this is a capability issue, and it is speculated that it is affected by the multi-linguist in some training sets.

\subsection{Other MLLMs as Judge}
We also use other mainstream MLLMs such as CogVLM \citep{wang2023cogvlm} and LLaVA \citep{liu2023llava} for judging, which hyper-parameters follow the response generation settings. Some issues arise when we use the basic ``Analyze-then-Judge'' setting, where they cannot follow the judging instructions very well. Therefore we do not put their results into the results of the text. Recently, we have also noticed the advent of GLM-4V \footnote{\url{https://open.bigmodel.cn/}}, which has shown good performance on some benchmarks and can receive long-term text and follow the instructions for judging. However, due to the time constraints, we have not completed tests on GLM-4V in our work.

\clearpage
\section{Additional Experimental Results}

\begin{figure*}[!t]
\vspace{-1em}
    \centering
    \includegraphics[width=.92\linewidth]{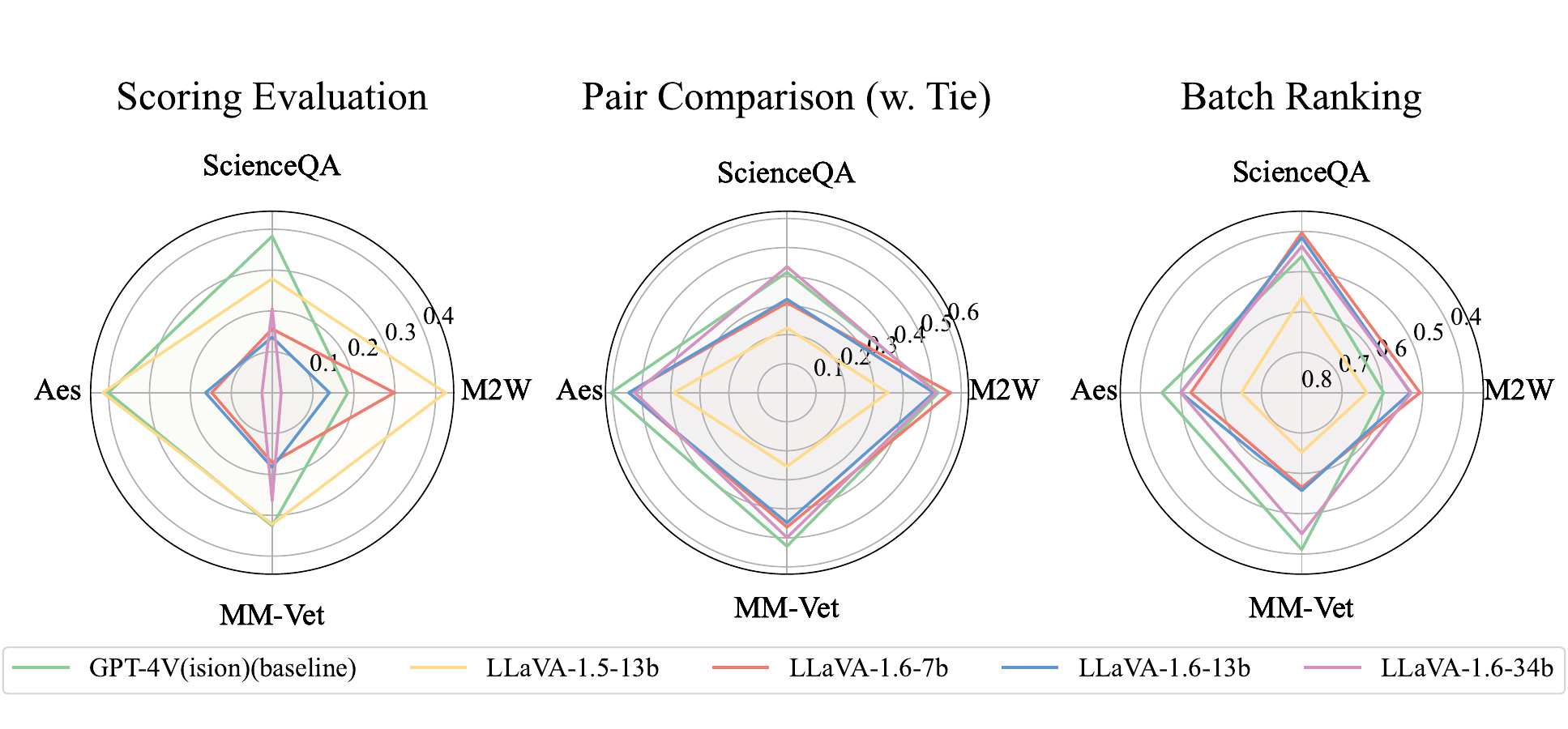}
      \vspace{-13pt}
    \caption{Comparative performance of different MLLMs across three judging settings in four newly added datasets, each is the average of three iterations. }
    \label{fig: rebuttal figure1}
      \vspace{-12pt}
\end{figure*}

\begin{figure*}[!t]
    \centering
    \includegraphics[width=.92\linewidth]{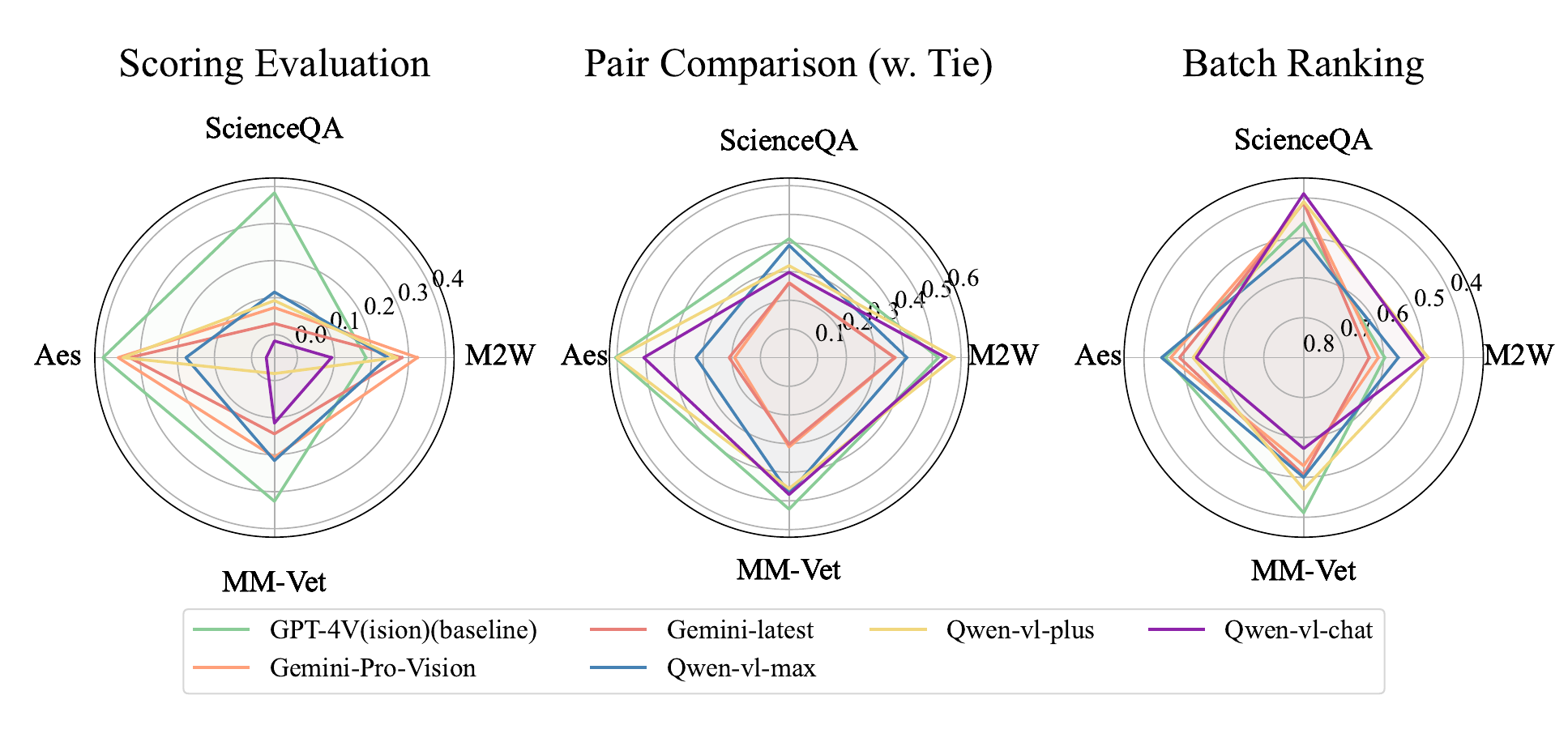}
      \vspace{-13pt}
    \caption{Comparative performance of different MLLMs across three judging settings in four newly added datasets, each is the average of three iterations. }
    \label{fig: rebuttal figure2}
      \vspace{-12pt}
\end{figure*}

\subsection{Full Results on Judging Performance\label{full}}
We provide full results of judging performance of different MLLMs in Table~\ref{tab: full result}. Comparative performance for four newly added datasets are shown in Figures~\ref{fig: rebuttal figure1} and \ref{fig: rebuttal figure2}.

In \textit{Scoring Evaluation}, all models demonstrated comparable performance levels on the original dataset presented in our study, with LLaVA-1.6-34b and Qwen-vl-max slightly outperforming others in Math, Chart, and Text tasks, yet none surpassing GPT-4V. Our analysis of Qwen-vl-max and Qwen-vl-plus revealed a propensity to assign higher scores, with 80\% of their ratings falling between 4-5 points, and a noticeable absence of 1-2 point scores. This inclination towards higher scores is more pronounced compared to other models. The LLaVA-1.6 series, although slightly better, also tends to award scores within the 3-5.
In Pair comparison, qwen-vl-plus and max performed better on certain datasets, distinguishing themselves from competitors. Notably, qwen-vl-max exhibited less positional bias than LLaVA models, which showed a strong predisposition to favor one position, typically rating `A' as better.
n Batch Ranking, the updated Gemini-Pro-Vision model outperforms others overall. Both Qwen and LLaVA series demonstrated that larger model sizes correlate with better outcomes, affirming a strong scaling law effect. Despite these findings, there remains a noticeable gap between these models and the top-performing GPT-4V, particularly concerning positional bias.

\begin{table*}[htbp]
\centering
\large
\renewcommand\arraystretch{1.2}

\caption{The overall performance of different MLLMs in judging, compared with human annotations on different datasets. We sample all the data three times and took the average to mitigate the casualty. \textit{w.} and \textit{w.o.} tie represents tie and non-tie situations respectively. We omit Gemini's results on the diffusion task for its challenges in processing AI-generated images. 
All presented data of Pearson similarity exhibit a $p$-value below 0.05, indicating a statistically significant level of confidence. Notice: Gemini-Pro$^{*}$ means Gemini-1.0-Pro-latest.
}

\label{tab: full result}
\resizebox{1\linewidth}{!}{
\begin{tabular}{ll|ccccccccccccccc}\toprule[1.5pt]
\textbf{Settings} & \textbf{MLLM}  & COCO  & C.C. & Diff. & Graphics & Math & Text & WIT & Chart & VisIT & CC-3M & M2W & SciQA & Aes & MM-Vet & Ave. \\ \midrule
\multirow{11}{*}{\textbf{Score} ($\uparrow$)} & CogVLM & 0.107 & -0.048 & 0.049 & -0.158 & 0.065 & 0.097 & -0.131 & -0.135 & 0.278 & 0.157 & - & - & - & - & 0.028 \\
 & GPT-4V & \textbf{0.454} & \textbf{0.507} & \textbf{0.458} & \textbf{0.645} & \textbf{0.606} & \textbf{0.624} & \textbf{0.579} & \textbf{0.645} & \textbf{0.620} & \textbf{0.431} & 0.185 & \textbf{0.383} & 0.401 & \textbf{0.326} & \textbf{0.490} \\
 & LLaVA-1.5-13b & 0.247 & 0.227 & 0.060 & 0.242 & 0.093 & 0.245 & 0.109 & 0.237 & 0.177 & 0.071 & \textbf{0.424} & 0.279 & \textbf{0.414} & 0.322 & 0.225 \\
 & LLaVA-1.6-7b & 0.300 & 0.243 & 0.058 & 0.200 & 0.090 & 0.193 & 0.044 & 0.085 & 0.228 & 0.026 & 0.299 & 0.156 & 0.148 & 0.171 & 0.160 \\
 & LLaVA-1.6-13b & 0.289 & 0.226 & -0.110 & 0.078 & 0.056 & 0.086 & 0.062 & 0.120 & 0.163 & 0.200 & 0.140 & 0.136 & 0.163 & 0.183 & 0.128 \\
 & LLaVA-1.6-34b & 0.285 & 0.251 & -0.012 & 0.262 & 0.238 & 0.258 & 0.151 & 0.318 & 0.198 & 0.109 & 0.022 & 0.206 & 0.025 & 0.265 & 0.184 \\
 & Gemini-Pro & 0.262 & 0.408 & - & 0.400 & 0.228 & 0.222 & 0.418 & 0.343 & 0.336 & 0.374 & 0.324 & 0.073 & 0.360 & 0.207 & 0.304 \\
 & Gemini-Pro$^{*}$ & 0.211 & 0.230 & 0.114 & 0.146 & 0.060 & 0.095 & 0.041 & 0.160 & 0.174 & 0.177 & 0.282 & 0.030 & 0.329 & 0.144 & 0.157 \\
 & Qwen-vl-max & 0.311 & 0.117 & 0.072 & 0.218 & 0.175 & 0.196 & 0.028 & 0.312 & 0.151 & 0.045 & 0.244 & 0.115 & 0.177 & 0.216 & 0.170 \\
 & Qwen-vl-plus & -0.050 & 0.195 & 0.019 & 0.126 & 0.106 & 0.161 & 0.151 & 0.089 & 0.128 & 0.106 & 0.268 & 0.092 & 0.347 & -0.019 & 0.123 \\
 & Qwen-vl-chat & -0.012 & -0.012 & 0.033 & -0.422 & 0.011 & -0.028 & 0.021 & 0.036 & -0.060 & 0.083 & 0.092 & -0.017 & -0.040 & 0.115 & -0.014 \\ \midrule
\multirow{11}{*}{\textbf{Pair w. Tie} ($\uparrow$)} & CogVLM & 0.548 & 0.409 & 0.562 & 0.613 & 0.412 & 0.250 & 0.273 & 0.262 & 0.324 & 0.433 & - & - & - & - & 0.409 \\
 & GPT-4V & \textbf{0.696} & \textbf{0.824} & \textbf{0.847} & \textbf{0.639} & 0.564 & \textbf{0.673} & \textbf{0.679} & \textbf{0.657} & 0.640 & \textbf{0.612} & 0.521 & 0.415 & \textbf{0.606} & \textbf{0.529} & \textbf{0.636} \\
 & LLaVA-1.5-13b & 0.273 & 0.478 & 0.286 & 0.273 & \textbf{0.657} & 0.510 & 0.369 & 0.383 & 0.456 & 0.484 & 0.347 & 0.223 & 0.389 & 0.254 & 0.384 \\
 & LLaVA-1.6-7b & 0.493 & 0.571 & 0.550 & 0.383 & 0.314 & 0.507 & 0.500 & 0.352 & 0.401 & 0.402 & 0.563 & 0.310 & 0.544 & 0.463 & 0.454 \\
 & LLaVA-1.6-13b & 0.493 & 0.586 & 0.590 & 0.333 & 0.339 & 0.507 & 0.587 & 0.296 & 0.454 & 0.459 & 0.506 & 0.322 & 0.545 & 0.448 & 0.462 \\
 & LLaVA-1.6-34b & 0.493 & 0.600 & 0.570 & 0.300 & 0.374 & 0.551 & 0.543 & 0.254 & 0.398 & 0.392 & 0.513 & \textbf{0.434} & 0.524 & 0.499 & 0.460 \\
 & Gemini-Pro & 0.616 & 0.787 & - & 0.650 & 0.436 & 0.664 & 0.605 & 0.500 & \textbf{0.660} & 0.560 & 0.370 & 0.262 & 0.190 & 0.312 & 0.509 \\
 & Gemini-Pro$^{*}$ & 0.273 & 0.273 & 0.240 & 0.324 & 0.237 & 0.275 & 0.136 & 0.377 & 0.232 & 0.294 & 0.368 & 0.260 & 0.209 & 0.303 & 0.272 \\
 & Qwen-vl-max & 0.403 & 0.464 & 0.372 & 0.494 & 0.438 & 0.500 & 0.533 & 0.479 & 0.421 & 0.421 & 0.411 & 0.392 & 0.325 & 0.474 & 0.438 \\
 & Qwen-vl-plus & 0.479 & 0.507 & 0.650 & 0.450 & 0.328 & 0.522 & 0.500 & 0.380 & 0.453 & 0.383 & \textbf{0.577} & 0.321 & 0.601 & 0.457 & 0.472 \\
 & Qwen-vl-chat & 0.493 & 0.486 & 0.480 & 0.311 & 0.248 & 0.406 & 0.543 & 0.310 & 0.332 & 0.292 & 0.547 & 0.298 & 0.507 & 0.478 & 0.409 \\  \midrule
\multirow{11}{*}{\textbf{Pair w.o. Tie} ($\uparrow$)} & CogVLM & 0.654 & 0.450 & 0.643 & 0.704 & 0.481 & 0.292 & 0.500 & 0.423 & 0.500 & 0.591 & - & - & - & - & 0.524 \\
 & GPT-4V & \textbf{0.804} & \textbf{0.870} & \textbf{0.922} & \textbf{0.807} & \textbf{0.801} & \textbf{0.805} & 0.734 & \textbf{0.849} & \textbf{0.761} & \textbf{0.703} & 0.699 & 0.647 & \textbf{0.755} & 0.659 & \textbf{0.773} \\
 & LLaVA-1.5-13b & 0.327 & 0.537 & 0.302 & 0.300 & 0.726 & 0.684 & 0.600 & 0.610 & 0.648 & 0.583 & 0.449 & 0.443 & 0.498 & 0.344 & 0.504 \\
 & LLaVA-1.6-7b & 0.593 & 0.597 & 0.618 & 0.434 & 0.468 & 0.636 & 0.561 & 0.471 & 0.436 & 0.466 & 0.633 & 0.621 & 0.568 & 0.705 & 0.558 \\
 & LLaVA-1.6-13b & 0.614 & 0.612 & 0.663 & 0.382 & 0.487 & 0.618 & 0.659 & 0.420 & 0.503 & 0.549 & 0.576 & 0.598 & 0.565 & 0.620 & 0.562 \\
 & LLaVA-1.6-34b & 0.607 & 0.824 & 0.855 & 0.402 & 0.587 & 0.750 & \textbf{0.758} & 0.381 & 0.503 & 0.564 & \textbf{0.712} & \textbf{0.679} & 0.694 & \textbf{0.762} & 0.648 \\
 & Gemini-Pro & 0.717 & 0.840 & - & 0.770 & 0.678 & 0.793 & 0.688 & 0.658 & 0.711 & 0.652 & 0.471 & 0.358 & 0.265 & 0.400 & 0.615 \\
 & Gemini-Pro$^{*}$ & 0.311 & 0.340 & 0.308 & 0.419 & 0.336 & 0.366 & 0.200 & 0.439 & 0.290 & 0.358 & 0.469 & 0.336 & 0.266 & 0.398 & 0.345 \\
 & Qwen-vl-max & 0.657 & 0.674 & 0.556 & 0.667 & 0.635 & 0.732 & 0.647 & 0.638 & 0.560 & 0.586 & 0.608 & 0.646 & 0.741 & 0.662 & 0.644 \\
 & Qwen-vl-plus & 0.596 & 0.556 & 0.771 & 0.554 & 0.463 & 0.735 & 0.575 & 0.535 & 0.521 & 0.510 & 0.659 & 0.612 & 0.627 & 0.659 & 0.598 \\
 & Qwen-vl-chat & 0.603 & 0.523 & 0.625 & 0.333 & 0.386 & 0.574 & 0.625 & 0.431 & 0.370 & 0.396 & 0.618 & 0.594 & 0.539 & 0.755 & 0.527 \\   \midrule
\multirow{10}{*}{\textbf{Batch} ($\downarrow$)} & GPT-4V & \textbf{0.318} & 0.353 & \textbf{0.070} & \textbf{0.385} & \textbf{0.348} & \textbf{0.319} & \textbf{0.290} & \textbf{0.347} & \textbf{0.300} & 0.402 & 0.597 & 0.462 & 0.453 & \textbf{0.411} & \textbf{0.361} \\
 & LLaVA-1.5-13b & 0.577 & 0.492 & 0.562 & 0.535 & 0.598 & 0.650 & 0.616 & 0.644 & 0.620 & 0.563 & 0.639 & 0.563 & 0.650 & 0.652 & 0.597 \\
 & LLaVA-1.6-7b & 0.575 & 0.538 & 0.618 & 0.462 & 0.601 & 0.598 & 0.564 & 0.679 & 0.586 & 0.503 & 0.507 & 0.403 & 0.525 & 0.565 & 0.552 \\
 & LLaVA-1.6-13b & 0.614 & 0.612 & 0.663 & 0.382 & 0.487 & 0.618 & 0.659 & 0.420 & 0.503 & 0.549 & 0.531 & 0.415 & 0.500 & 0.557 & 0.536 \\
 & LLaVA-1.6-34b & 0.449 & 0.411 & 0.500 & 0.561 & 0.575 & 0.544 & 0.483 & 0.552 & 0.542 & 0.479 & 0.529 & 0.437 & 0.500 & 0.450 & 0.501 \\
 & Gemini-Pro & 0.287 & \textbf{0.299} & - & 0.473 & 0.462 & 0.430 & 0.344 & 0.520 & 0.426 & \textbf{0.357} & 0.613 & 0.412 & 0.467 & 0.529 & 0.432 \\
 & Gemini-Pro$^{*}$ & 0.378 & 0.370 & - & 0.572 & 0.508 & 0.452 & 0.417 & 0.572 & 0.492 & 0.434 & 0.636 & 0.412 & 0.489 & 0.506 & 0.480 \\
 & Qwen-vl-max & 0.477 & 0.407 & 0.500 & 0.480 & 0.507 & 0.515 & 0.493 & 0.539 & 0.468 & 0.407 & 0.563 & 0.503 & \textbf{0.444} & 0.500 & 0.486 \\
 & Qwen-vl-plus & 0.640 & 0.616 & 0.500 & 0.666 & 0.644 & 0.634 & 0.592 & 0.747 & 0.671 & 0.540 & \textbf{0.488} & 0.409 & 0.523 & 0.470 & 0.581 \\
 & Qwen-vl-chat & 0.733 & 0.701 & 0.500 & 0.669 & 0.638 & 0.554 & 0.638 & 0.723 & 0.687 & 0.668 & 0.500 & \textbf{0.389} & 0.531 & 0.572 & 0.607\\
 \bottomrule[1.5pt]
\end{tabular}}
  \vspace{-10pt}
\end{table*}

\subsection{Judging Results on Sequential Images}
\label{momentos}
We incorporated the sequential image dataset Mementos, comprising picture sequences, to expand our MLLM-as-a-Judge framework into the video domain in a pioneering effort. Each sequence, featuring over four images, draws from daily life, comics, and robotics. For data generation in Step 3, we utilized GPT-4V, Qwen-VL-Max, Qwen-VL-Plus, and Video-LLM Chatunivi, obtaining 100 image-text pairs for batch evaluations, 381 for scoring, and 560 for pair comparisons. Beyond analyzing GPT-4V and Qwen-vl-max, we explored Video-LLM's judging capabilities, specifically testing it with ChatUnivi. 

As illustrated in Table \ref{tab: sequence} for \textit{Batch Evaluation}, \textit{Pair Comparison}, and \textit{Score Evaluation} respectively, our findings indicate that GPT-4V significantly outperforms other models on sequential data. Despite the high-quality responses generated by the Video-LLM ChatUnivi we evaluated, it fundamentally lacks the judging capability and consistency.

\begin{table}[htbp]
\centering
\renewcommand\arraystretch{1.1}
\caption{Judging performance on image sequence dataset Mementos. }
\label{tab: sequence}
\resizebox{0.45\linewidth}{!}{
\begin{tabular}{ll|cc|cc}
\toprule[1.5pt]
\multirow{2}{*}{MLLM} & \textbf{Score ($\uparrow$)} & \multicolumn{2}{c|}{\textbf{Pair ($\uparrow$)}} & \textbf{Batch ($\downarrow$)} \\
 &   Pearson & w. Tie & w.o. Tie & Edit Dis. \\ \midrule
GPT-4V & \textbf{0.361} & \textbf{0.721} & \textbf{0.836} & \textbf{0.411}  \\
ChatUnivi & -0.094 & 0.158 & 0.168 & 0.556  \\
Qwen-vl-plus & 0.115 & 0.426 & 0.482 & 0.5  \\
Qwen-vl-max & 0.046 & 0.446 & 0.531 & 0.63 \\
 \bottomrule[1.5pt]
\end{tabular}}
  \vspace{-5pt}
\end{table}

\subsection{Preliminary Experiment}
\noindent\textbf{Human Agreement on GPT-4V Output Mode.}
\label{Human Agreement on GPT-4V Output Mode}
The recently introduced `Json Mode'\footnote{\url{https://openai.com/blog/new-models-and-developer-products-announced-at-devday}} in GPT-4V represents a significant advancement, particularly in structuring outputs in JSON format while restricting token usage. This mode has been observed to regularize responses, a feature particularly advantageous when dealing with structured data. However, this structuring tends to compartmentalize responses, potentially leading to a loss in the natural flow and contextual linkage typically inherent in human-like responses. This segmentation might inadvertently affect the readability and perceived coherence of the generated text.

To quantitatively assess the impact of Json Mode on output quality and its alignment with human preferences, we meticulously construct a test set. This set comprises 50 data instances, randomly selected from three distinct datasets used for evaluation purposes. The objective is to discern human evaluators' predilection for the outputs generated in Json Mode by GPT-4V.

\begin{figure}[h]
    \centering
    \includegraphics[width=0.4\linewidth]{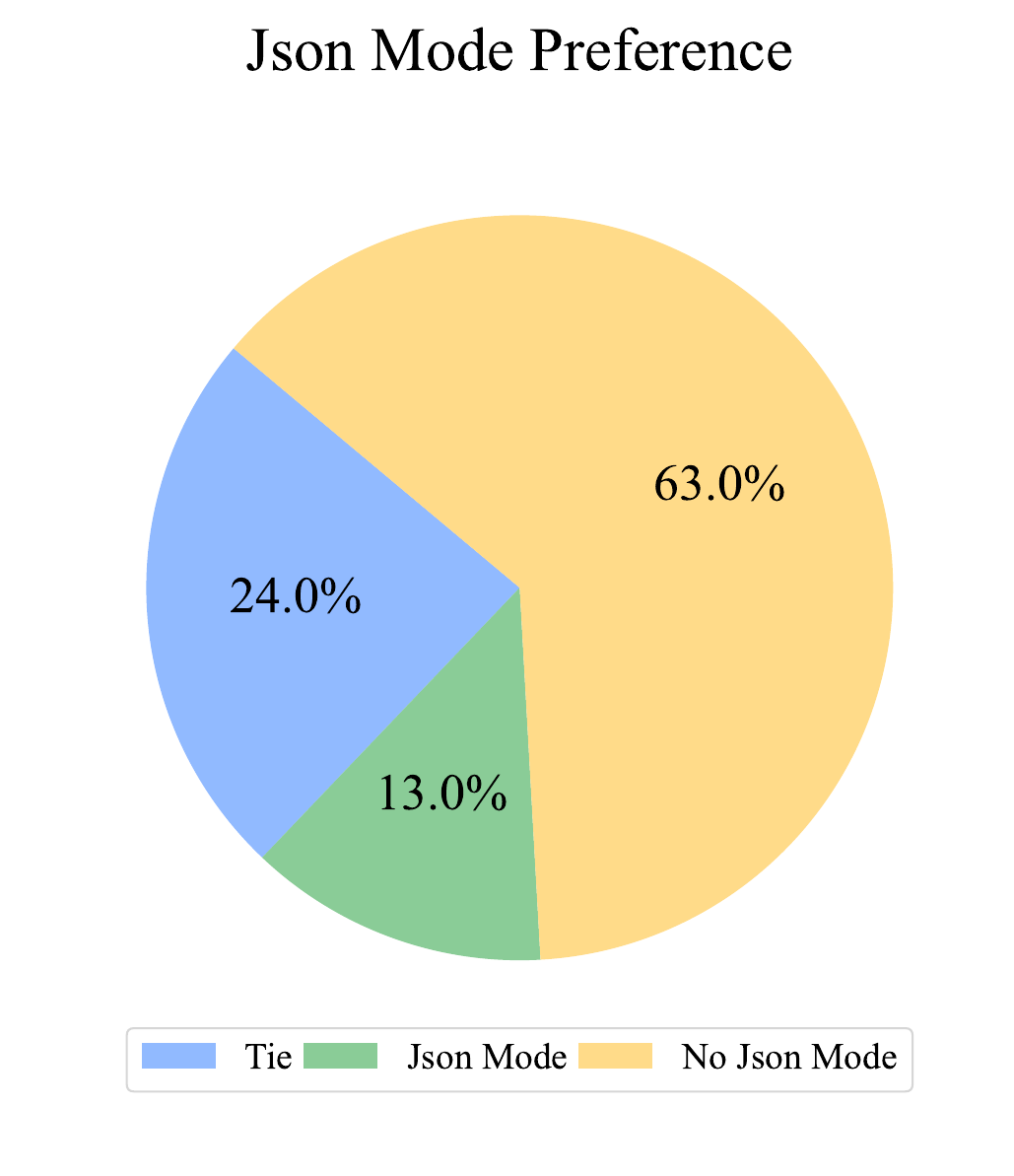}
        \vspace{-15pt}
    \caption{Json Mode Preference Analysis.}
\vspace{-5pt}
    \label{fig:jsonmode}
\end{figure}

For a comprehensive analysis, we engage three annotators, each responsible for labeling the data. Their assessments aim to discern the balance between structured, JSON-formatted responses and the inherently fluid nature of human judgment and preference in textual content, as shown in Figure~\ref{fig:jsonmode}.

\noindent\textbf{Human Agreement Bias Checking}
Acknowledging the inherent variability in human annotations, we embark on an empirical study involving ten annotators to ascertain the reliability of derived statistical patterns, notwithstanding the subjective nature of human judgment. This study aims to mitigate the individual biases that might skew the evaluation of GPT-4's outputs. A dataset comprising 50 entries, processed using the GPT-4 pair comparison setting, serves as the foundation for this investigation.

\begin{figure}[htbp]
    \centering
    \subfigure[The distribution of Human Annotators' ratings for the data.]{
        \includegraphics[width=1\linewidth]{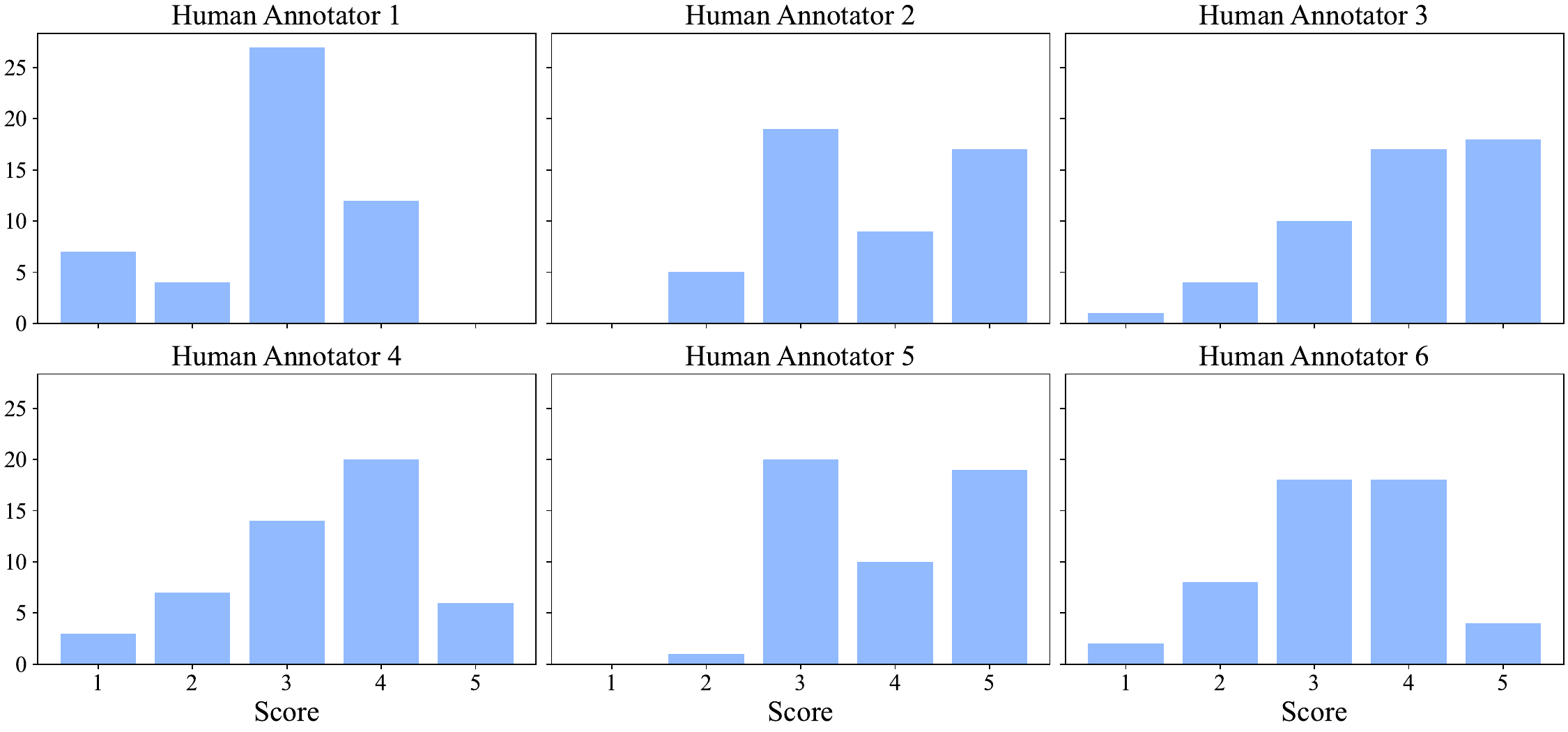}}\\
    \subfigure[Human Labeling and Agreement Bias Checking.]{
        \includegraphics[width=.6\linewidth]{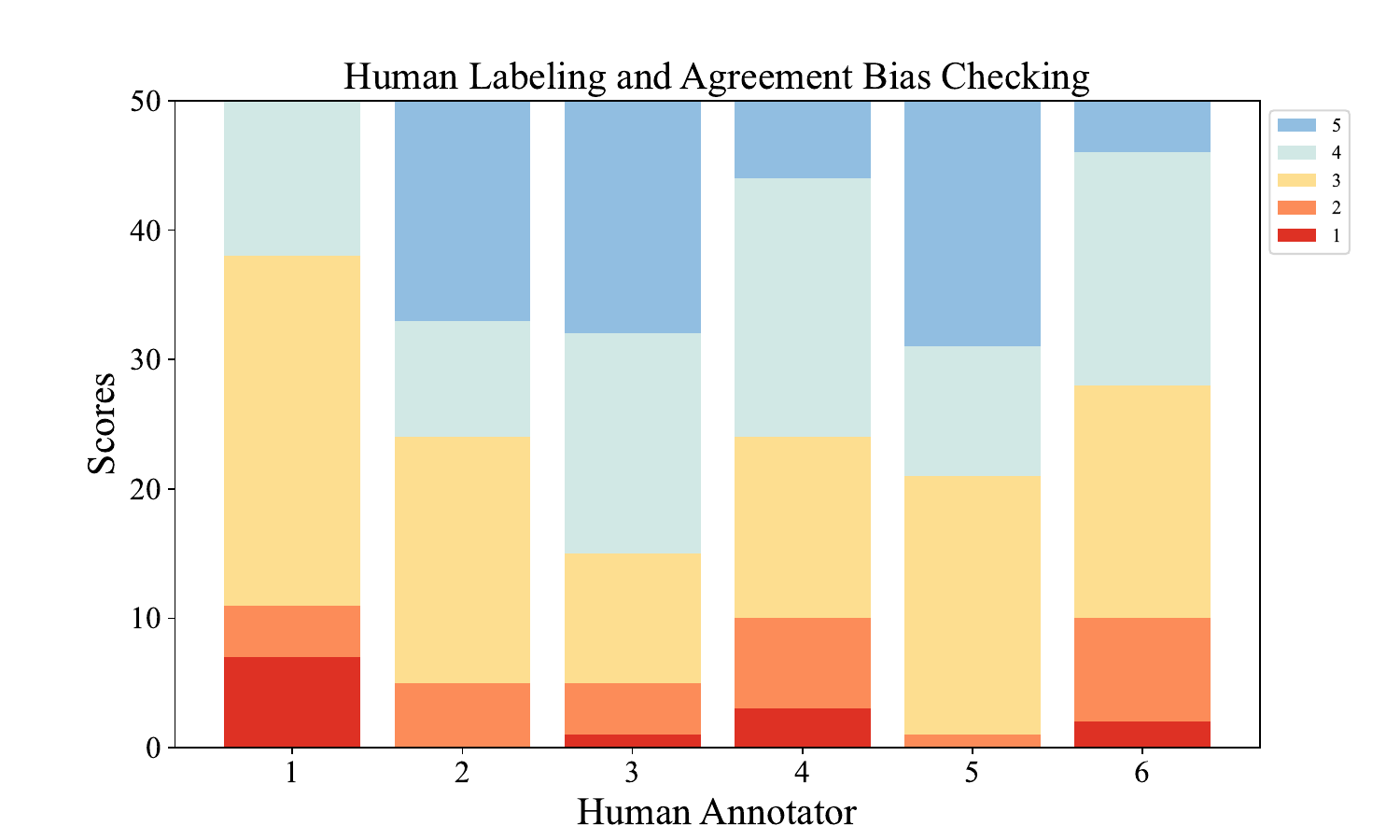}}
    \caption{Human Labeling and Agreement Bias.}
    \label{fig:human agreement bias}
\end{figure}

The results, detailed in Figure \ref{fig:human agreement bias}, underscore a noteworthy observation: while the annotators exhibit minimal variance in determining the correctness of GPT-4's judgments, a discernible divergence emerged in the scoring of analytical responses. This divergence presumably stems from individual perceptual differences and inherent biases. However, it's crucial to note that these discrepancies in scoring did not significantly compromise the overall integrity of the annotations.

A remarkable consensus is observed in the labeling of hallucinations. The employment of a meticulously defined decision tree for identifying hallucinations ensures a high degree of uniformity across the annotations. This structured approach substantially minimizes errors, underscoring the effectiveness of well-defined criteria in achieving consistent and reliable annotations across different individuals.

\subsection{Length Distribution on MLLM Judgments Analysis}
\begin{figure}[ht]
    \centering
    \includegraphics[width=\linewidth]{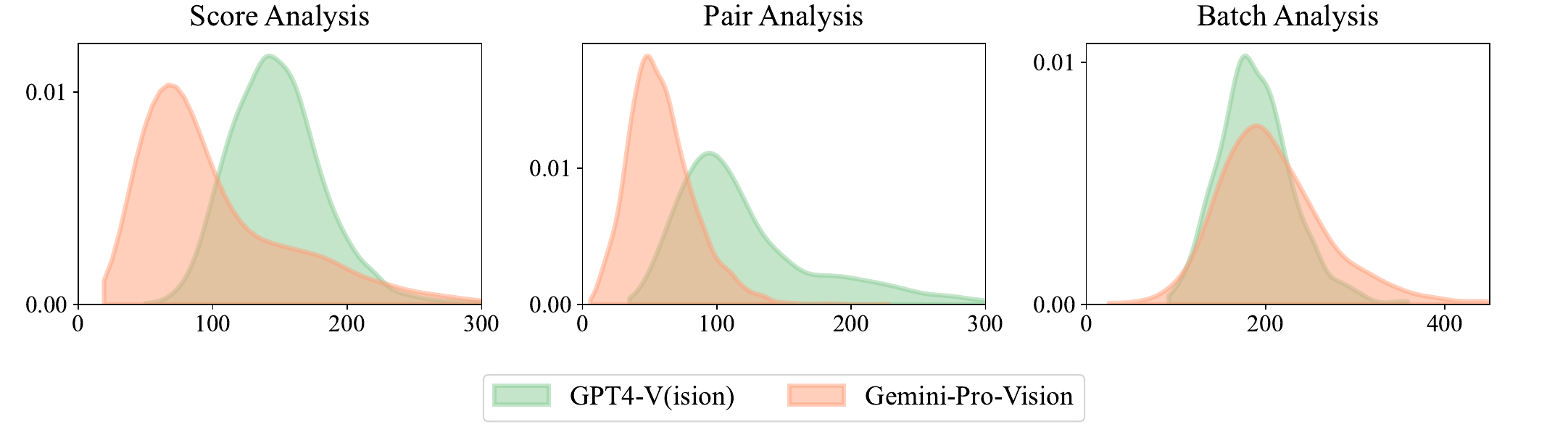}
    \vspace{-15pt}
    \caption{Length distribution in analysis collections.}
    \vspace{-10pt}
    \label{fig: length_distribution}
\end{figure}

\begin{figure}[h]
    \centering
    \includegraphics[width=0.9\linewidth]{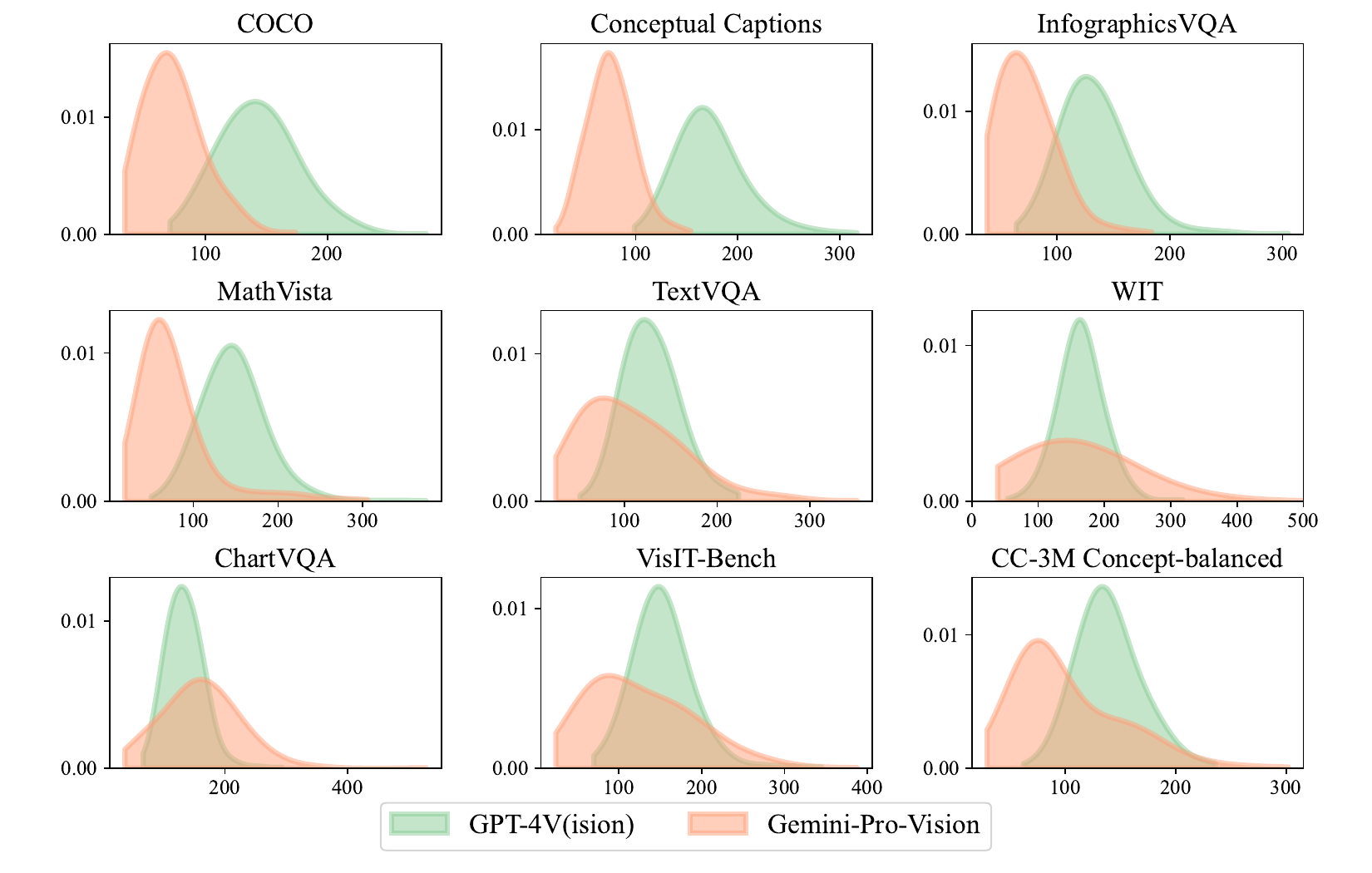}
    \vspace{-15pt}
    \caption{Response length distribution in Scoring Evaluation. The horizontal axis represents length, and the vertical axis represents density.}
    \vspace{-10pt}
    \label{fig: detailed score length distribution}
\end{figure}

\begin{figure}[htbp]
    \centering
    \includegraphics[width=0.9\linewidth]{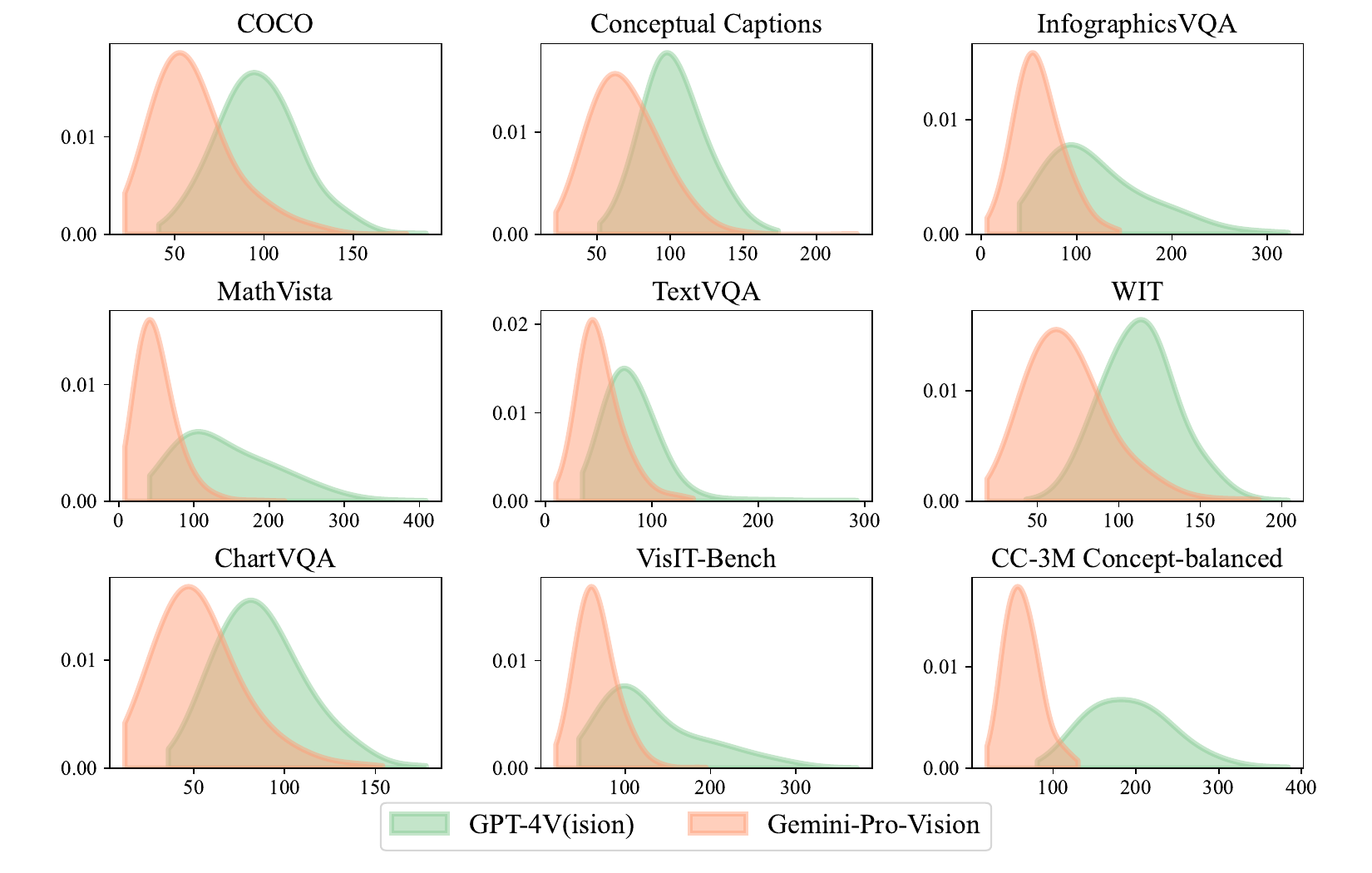}
    \vspace{-15pt}
    \caption{Response length distribution in Pair Comparison. }
    \vspace{-10pt}
    \label{fig: detailed pair length distribution}
\end{figure}

\begin{figure}[htbp]
    \centering
    \includegraphics[width=0.9\linewidth]{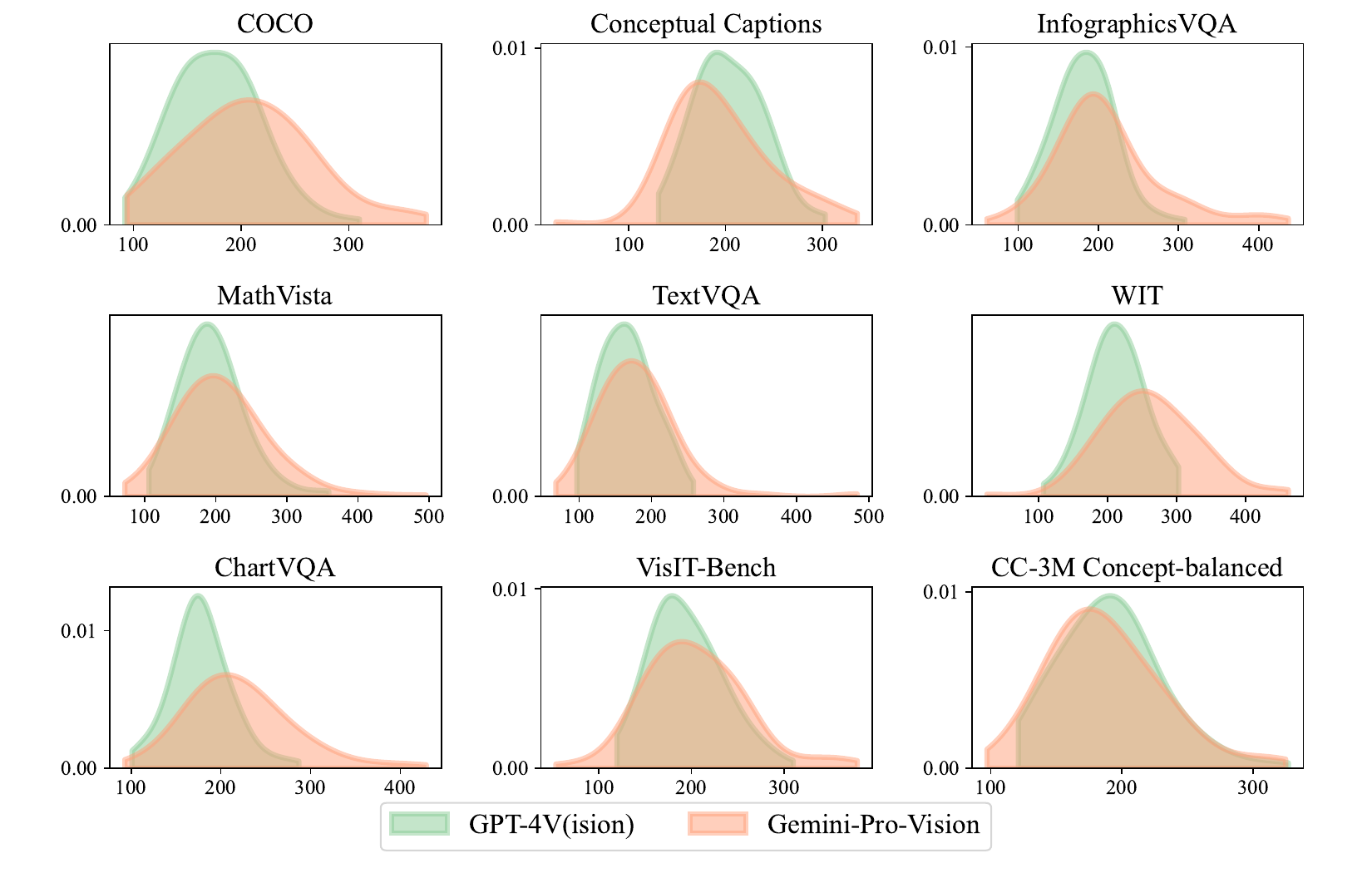}
     \vspace{-10pt}
    \caption{Response length distribution in Batch Ranking. }
     \vspace{-10pt}
    \label{fig: detailed batch length distribution}
\end{figure}

In our analysis, we have included length distribution diagrams that showcase the differences in the responses provided by GPT-4V and Gemini during their judgment tasks as illustrated in Figure~\ref{fig: length_distribution}. These diagrams reveal that GPT-4V typically generates longer responses than Gemini in both \textit{Scoring Evaluation} (Figure~\ref{fig: detailed score length distribution}) and \textit{Pair Comparison} (Figure~\ref{fig: detailed pair length distribution}), whereas in the batch task (Figure~\ref{fig: detailed batch length distribution}), the output lengths from both models are comparatively similar.


\newpage
\subsection{Results on Human Scoring and Ego Bias}
\begin{figure}[ht]
    \centering
    \includegraphics[width=0.5\linewidth]{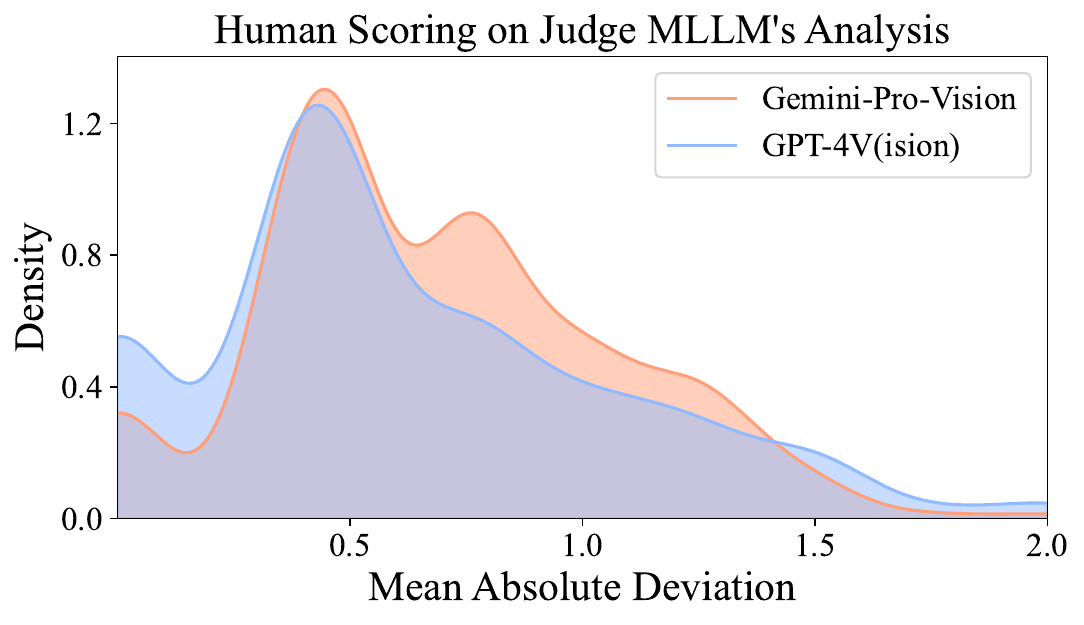}
    \vspace{-8pt}
    \caption{MAD of Human Scoring on MLLM Judgments Analysis.}
    \label{fig:MAD}
     
\end{figure}

We employ the Mean Absolute Deviation (MAD) metric to assess the consistency of MLLM judging quality across multiple responses to a single image-instruction pair, as shown in \ref{fig:MAD}.

\begin{figure*}[h]
    \centering
    \includegraphics[width=\linewidth]{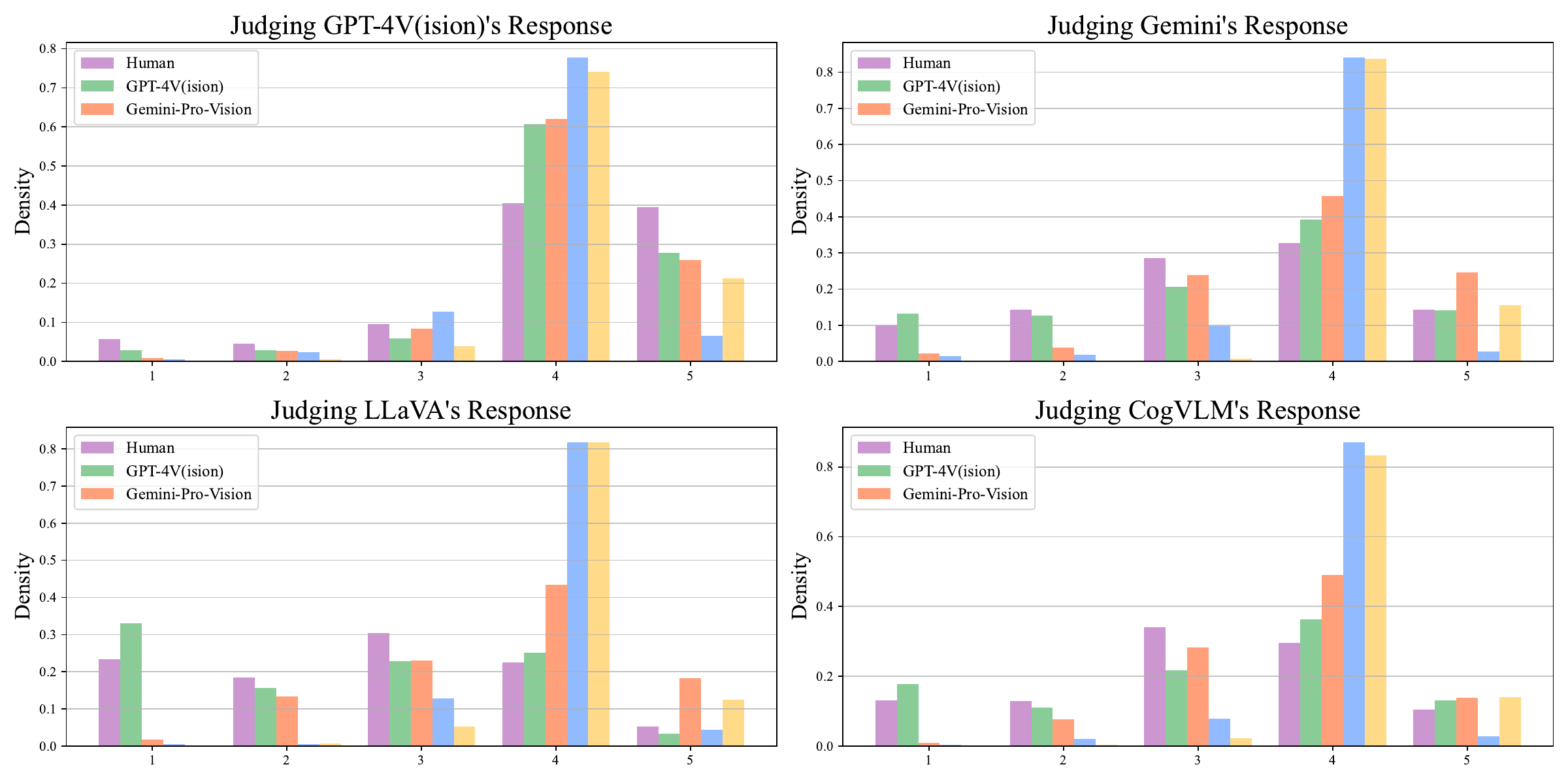}
    \vspace{-15pt}
    \caption{Scoring Density of Different MLLMs in Judging.}
    \label{fig: Ego Bias}
\end{figure*}

\begin{figure*}[h]
    \centering
    \includegraphics[width=\linewidth]{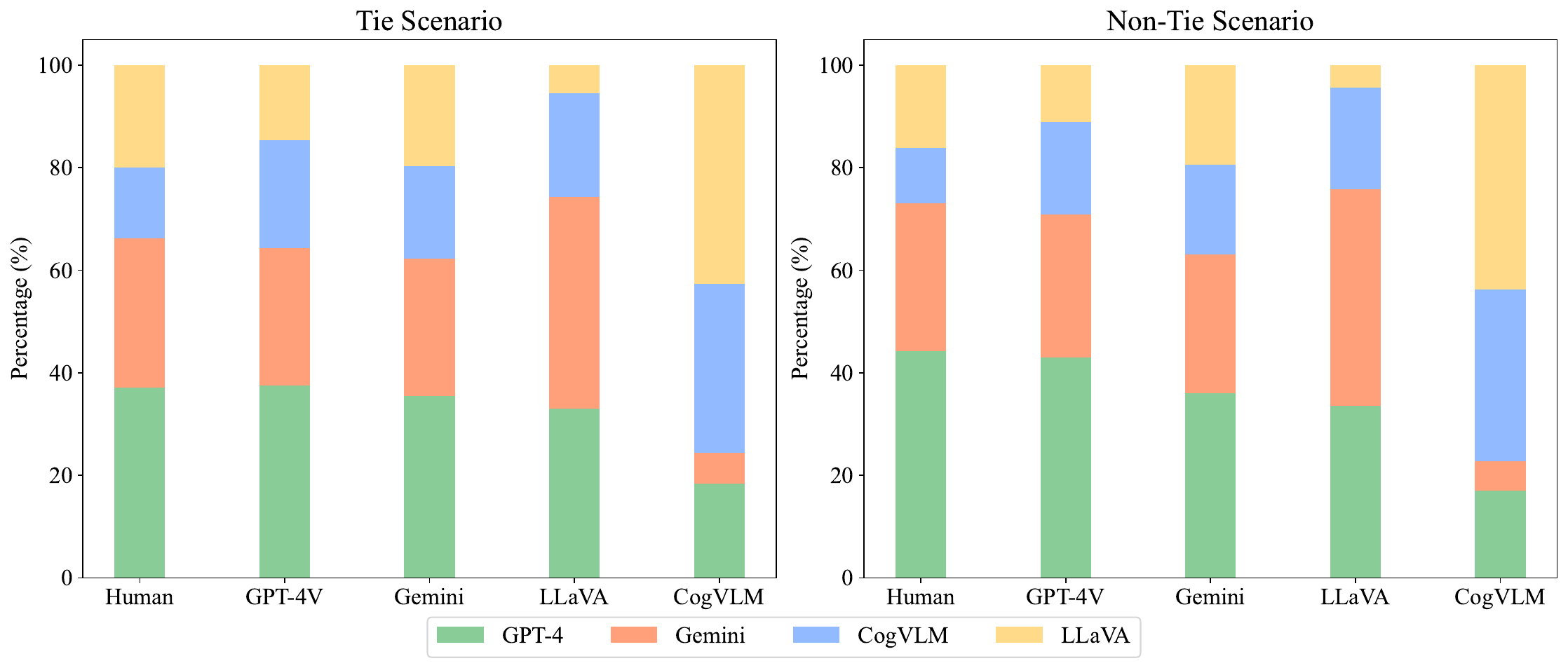}
    \vspace{-15pt}
    \caption{The proportion of different responses chosen by humans and different MLLMs in Tie Scenario and Non-Tie Scenario.}
    \label{fig: Ego Bias on pair}
\end{figure*}

The Egocentric Bias of different models are shown in Figures \ref{fig: Ego Bias} and \ref{fig: Ego Bias on pair}.

\newpage

\section{Human Labeling and Agreement Collection}
\label{Human Labeling and Agreement Collection}
The annotation is conducted by 6 authors of this paper independently. As acknowledged, the diversity of annotators plays a crucial role in reducing bias and enhancing the reliability of the benchmark. These annotators have knowledge in this domain, with different genders, ages, and educational backgrounds. To ensure the annotators can proficiently mark the data, we provide them with detailed tutorials, teaching them how to evaluate model responses more objectively. Specifically, they are required to give judgments without considering answer lengths, and certain names or positions of the response. Besides, we implement cross-validation between different annotators and conduct continuous monitoring to ensure they are maintaining objectivity and fairness.

\begin{figure*}[htbp]
\centering
\includegraphics[width=1\linewidth]{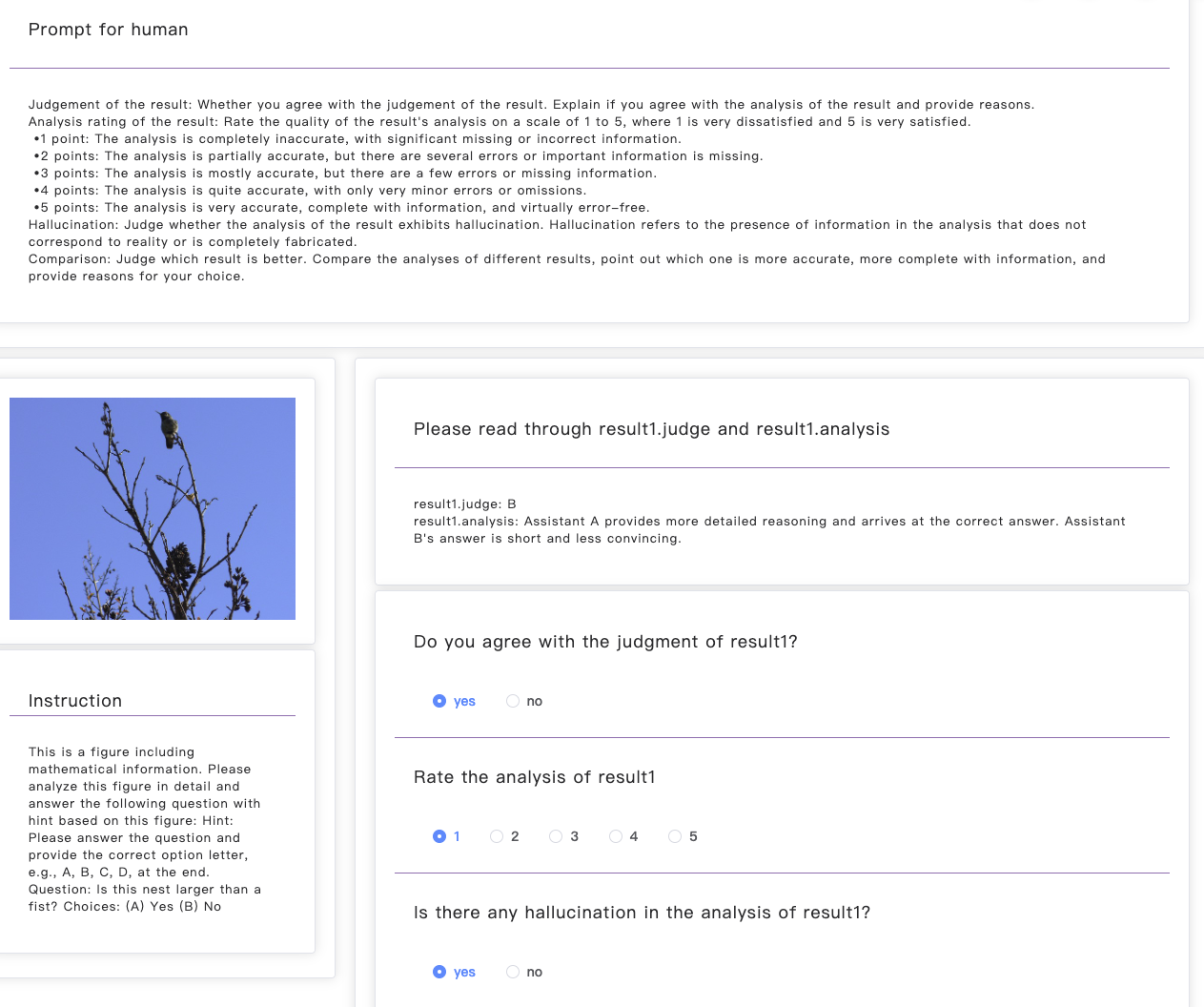}
\vspace{-10pt}
\caption{Human agreement}
\vspace{-10pt}
\label{human agreement}
\end{figure*}

\begin{figure*}[htbp]
\centering
\includegraphics[width=1\linewidth]{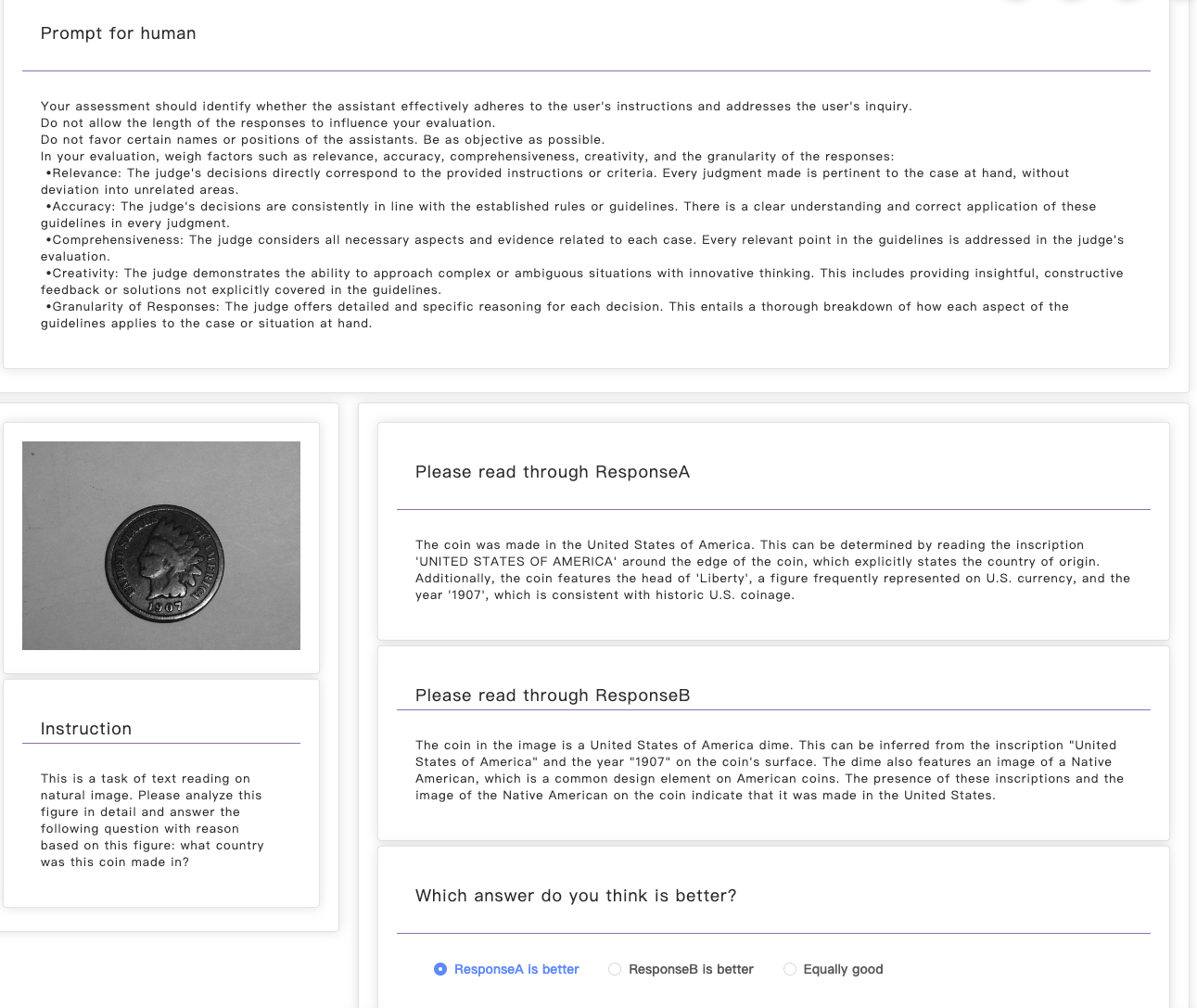}
\vspace{-10pt}
\caption{Human labeling}
\vspace{-10pt}
\label{human labeling}
\end{figure*}

In the Human agreement experiment performed by humans on MLLM Judge, the prompt we give humans is presented in Figure \ref{human agreement} and Figure \ref{human labeling}.

\newpage
\begin{center}
\begin{tcolorbox}[colback=gray!00,
                  colframe=black,
                  width=17.2cm,
                  arc=1.5mm, auto outer arc,
                  breakable,
                  left=0.9mm, right=0.9mm,
                  boxrule=0.9pt,
                  title = {Prompts for Human Agreement Experiment}
                 ]
Your assessment should identify whether the assistant effectively adheres to the user's instructions and addresses the user's inquiry. 

Do not allow the length of the responses to influence your evaluation. 

Do not favor certain names or positions of the assistants. Be as objective as possible.

In your evaluation, weigh factors such as relevance, accuracy, comprehensiveness, creativity, and the granularity of the responses:

Relevance: The judge's decisions directly correspond to the provided instructions or criteria. Every judgment made is pertinent to the case at hand, without deviation into unrelated areas.

Accuracy: The judge's decisions are consistently in line with the established rules or guidelines. There is a clear understanding and correct application of these guidelines in every judgment.

Comprehensiveness: The judge considers all necessary aspects and evidence related to each case. Every relevant point in the guidelines is addressed in the judge's evaluation.

Creativity: The judge demonstrates the ability to approach complex or ambiguous situations with innovative thinking. This includes providing insightful, constructive feedback or solutions not explicitly covered in the guidelines.

Granularity of Responses: The judge offers detailed and specific reasoning for each decision. This entails a thorough breakdown of how each aspect of the guidelines applies to the case or situation at hand.
\end{tcolorbox}
\end{center}

\clearpage

\section{Prompt Templates}
\label{Prompt templates}
We first query Judge VLM to get their responses by the following prompts.
\begin{center}
\begin{tcolorbox}[colback=gray!00,
                  colframe=black,
                  width=17.2cm,
                  arc=1.5mm, auto outer arc,
                  breakable,
                  left=0.9mm, right=0.9mm,
                  boxrule=0.9pt,
                  title = {Query prompts of MLLMs in judging.}
                 ]
You are a helpful assistant proficient in analyzing vision reasoning problems.\\
~[The Start of User Instruction]\\
~\{item['instruction']\}\\
~[The End of User Instruction]\\
Please provide a detailed explanation for your response.
\end{tcolorbox}
\end{center}

Following \citet{chiang2023closer} and \citet{li2024leveraging}, we have designed prompts and presented the prompt template of VLM's operation including score, pair comparison, and batch ranking judgments in a prompt template as \textit{system prompt, instruction, criteria, noticement, and desired output form}:

\begin{center}
\begin{tcolorbox}[colback=gray!00,
                  colframe=black,
                  width=17.2cm,
                  arc=1.5mm, auto outer arc,
                  breakable,
                  left=0.9mm, right=0.9mm,
                  boxrule=0.9pt,
                  title = {Template prompts of scoring evaluation}
                 ]
\textbf{(System Prompt)} \\
You are a helpful assistant proficient in analyzing vision reasoning problems. \\
\textbf{(Instruction)} \\
Please examine the provided image attentively and serve as an unbiased judge in assessing the quality of the response from an AI assistants regarding the instruction. You will receive a single response from the assistant to user's instruction. \\
\textbf{(Noticement)} \\
Your assessment should identify whether the assistant effectively adheres to the user's instructions and addresses the user's inquiry. \\
In your evaluation, weigh factors such as relevance, accuracy, comprehensiveness, creativity, and the granularity of the responses. \\
Do not allow the length of the responses to influence your evaluation. \\
Do not favor certain names or positions of the assistants. Be as objective as possible.\\
\textbf{(Criteria)} \\
Use scores to show the quality of the response. Here is the detailed scoring rubric for evaluating the quality of responses from AI assistants:\\
Poor (1): The response significantly deviates from the user's instruction and fails to address the query effectively. It shows a lack of relevance, accuracy, and comprehensiveness. Creativity and granularity are absent or poorly executed.\\
Fair (2): The response addresses the user's instruction partially, with evident shortcomings in relevance, accuracy, or comprehensiveness. It lacks depth in creativity and granularity, indicating a superficial understanding of the user's inquiry.\\
Average (3): The response adequately addresses the user's instruction, showing a fair level of relevance, accuracy, and comprehensiveness. It reflects a basic level of creativity and granularity but may lack sophistication or depth in fully capturing the user's inquiry.\\
Good (4): The response is well-aligned with the user's instruction, demonstrating a high degree of relevance, accuracy, and comprehensiveness. It shows creativity and a nuanced understanding of the topic, with a detailed granularity that enhances the response quality.\\
Excellent (5): The response perfectly adheres to the user's instruction, excelling in relevance, accuracy, comprehensiveness, creativity, and granularity. It provides an insightful, detailed, and thorough answer, indicating a deep and nuanced understanding of the user's inquiry.\\
\textbf{(Desired Output Format)} \\
Use "[[1]]", "[[2]]", "[[3]]", "[[4]]", "[[5]]" to indicate your evaluate score in the key `Judgement'.\\
~[The Start of User Instruction]\\
~\{item[`instruction']\}\\
~[The End of User Instruction]\\
~[The Start of Assistant’s Answer]\\
~\{item[`answer']\}\\
~[The End of Assistant’s Answer]
\end{tcolorbox}
\end{center}

\clearpage

\begin{center}
\begin{tcolorbox}[colback=gray!00,
                  colframe=black,
                  width=17.2cm,
                  arc=1.5mm, auto outer arc,
                  breakable,
                  left=0.9mm, right=0.9mm,
                  boxrule=0.9pt,
                  title = {Template prompts of pair comparison}
                 ]
\textbf{(System Prompt)} \\
You are a helpful assistant proficient in analyzing vision reasoning problems. \\
\textbf{(Instruction)} \\
Please examine the provided image attentively and serve as an unbiased judge in assessing the quality of responses from two AI assistants regarding the user's question shown beneath the image. \\
\textbf{(Noticement)} \\
Your assessment should identify the assistant that more effectively adheres to the user's instruction and aptly addresses the user's inquiry. \\
In your evaluation, weigh factors such as relevance, accuracy, comprehensiveness, creativity, and the granularity of the responses. \\
Avoid any position biases and ensure that the order in which the responses were presented does not influence your decision. \\
Do not allow the length of the responses to influence your evaluation. \\
Do not favor certain names of the assistants. Be as objective as possible.\\
Present your verdict in a JSON format, with the key 'analysis' for a short reason of your judgement and the key 'judgment' to indicate your decision: use ``[[A]]" if assistant A prevails, "[[B]]" if assistant B does, and "[[C]]" for a tie.\\
\textbf{(Desired Output Format)} \\
~[The Start of User Instruction].\\
~\{item[`instruction']\}\\
~[The End of User Instruction]\\
~[The Start of Assistant A’s Answer]\\
~\{item[`answer1'][`answer']\}\\
~[The End of Assistant A’s Answer]\\
~[The Start of Assistant B’s Answer]\\
~\{item[`answer2'][`answer']\}\\
~[The End of Assistant B’s Answer]
\end{tcolorbox}
\end{center}

\begin{center}
\begin{tcolorbox}[colback=gray!00,
                  colframe=black,
                  width=17.2cm,
                  arc=1.5mm, auto outer arc,
                  breakable,
                  left=0.9mm, right=0.9mm,
                  boxrule=0.9pt,
                  title = {Template prompts of batch ranking}
                 ]
\textbf{(System Prompt)} \\
You are a helpful assistant proficient in analyzing vision reasoning problems. \\
\textbf{(Instruction)} \\
Please serve as an unbiased judge in assessing the quality of the responses from AI assistants regarding the user's instruction and a figure. \\
You will be presented with several responses from different assistants to the same user instruction. 
Your task is to assess and compare these responses based on how effectively they adhere to the user's original instruction and how aptly they address the user's inquiry. 
After your assessment and comparison, you should RANK the responses from best to worst as the following template. If Assistant A is the best response, Assistant D is the worst response, you should output like [[A]], [[B]], [[C]], [[D]]"\\
\textbf{(Noticement)} \\
Your assessment should identify whether the assistant effectively adheres to the user's instructions and addresses the user's inquiry. \\
In your evaluation, weigh factors such as relevance, accuracy, comprehensiveness, creativity, and the granularity of the responses. \\
Do not allow the length of the responses to influence your evaluation. \\
Do not favor certain names or positions of the assistants. Be as objective as possible.\\
Indicate your final rank in the key 'Judgement'.\\
\textbf{(Desired Output Format)} \\
~[The Start of User Instruction].\\
~\{item[`instruction']\}\\
~[The End of User Instruction]
\end{tcolorbox}
\end{center}

\section{Case Study}
\label{Case Study}
We list several case studies. Assistant A, B, C, D represent LlaVA, CogVLM, GPT-4V and Gemini, respectively. Judge VLM 1 and 2 represent Gemini and GPT-4V, respectively. For pair comparison, the judge result ``C" means a tie.

\begin{tcolorbox}[colback=gray!00,
                  colframe=black,
                  width=17.2cm,
                  arc=1.5mm, auto outer arc,
                  breakable,
                  left=0.9mm, right=0.9mm,
                  boxrule=0.9pt, colbacktitle = black!65!black,
                  title = {An example of score comparison}
                 ]
    \begin{center}
    \includegraphics[width = 0.4\linewidth]{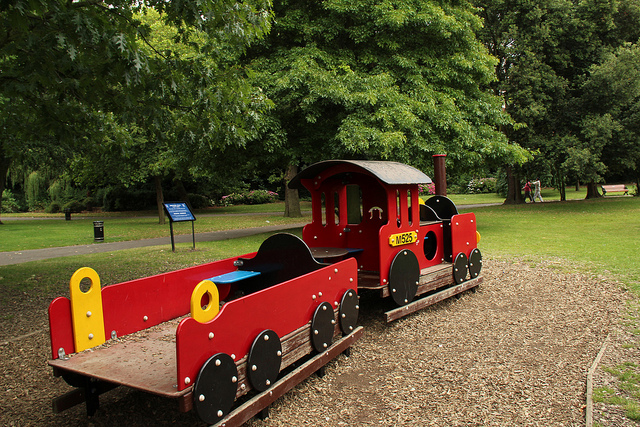}
    \end{center}
\colorbox{orange!30}{\textbf{Instruction:} Please analyze this figure in detail and answer the following question based on this figure: Who was the} \colorbox{orange!30}{train built for?} \\
\vspace{2pt}\\
\textbf{answer1.name:} Assistant A\\
\textbf{answer1.answer}: The train was built for children to play with in a park. The image shows a small red and black toy train parked on a dirt surface in a park, surrounded by trees. The train appears to be designed for children to ride on and enjoy, making it an ideal attraction for families visiting the park.\\
\textbf{result1-1.name}: Judge VLM 1\\
\textbf{result1-1.analysis}:  {The assistant's response is relevant to the user's instruction and addresses the user's inquiry. The response is accurate and comprehensive, providing a clear and concise answer to the question. The response is also creative and uses detailed granularity to enhance the quality of the response. \colorbox{orange!30}{Judgement: 5}}\\
\textbf{result1-2.name:} Judge VLM 2\\
\textbf{result1-2.analysis:} {The response provided by the AI assistant correctly identifies the purpose of the toy train in the image, stating that it was built for children to play with in a park. The description of the train as small, red and black, and parked on a dirt surface surrounded by trees is accurate based on the visual evidence in the image. Furthermore, the assistant correctly deduces that the train is designed for children to ride on and asserts that it is an ideal attraction for families, which is a logical conclusion given the context of the setting. The response is relevant to the user's instruction and demonstrates accuracy and a reasonable level of comprehensiveness. There is a minor display of creativity in the assistant's description of the train being an ideal attraction for families. \colorbox{orange!30}{Judgement: 4}}
\end{tcolorbox}

\begin{tcolorbox}[colback=gray!00,
                  colframe=black,
                  width=17.2cm,
                  arc=1.5mm, auto outer arc,
                  breakable,
                  left=0.9mm, right=0.9mm,
                  boxrule=0.9pt, colbacktitle = black!65!black,
                  title = {An example of pair comparison}
                 ]
    \begin{center}
    \includegraphics[width = 0.5\linewidth]{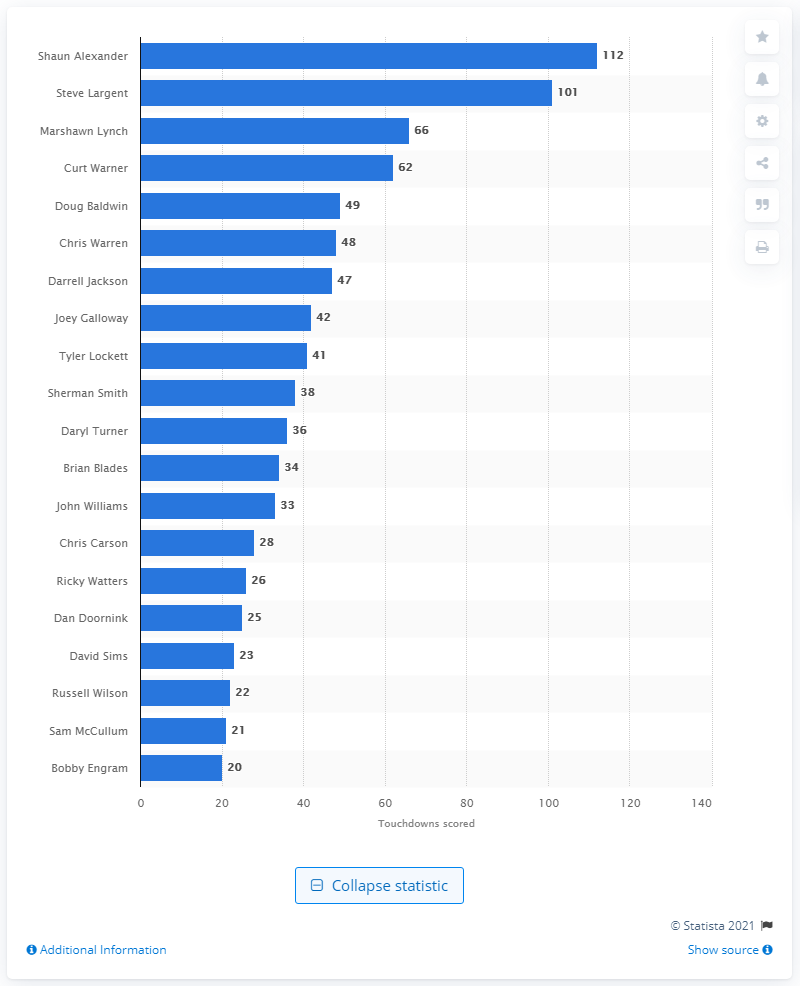}
    \end{center}
\colorbox{orange!30}{\textbf{Instruction:} This is a task of reasoning on chart data. Please analyze this figure in detail and answer the following } \colorbox{orange!30}{question with reason based on this figure. Who is the career touchdown leader of the Seattle Seahawks?} \\
\vspace{2pt}\\
\textbf{answer1.name:} Assistant A\\
\textbf{answer1.answer}: The career touchdown leader of the Seattle Seahawks is Shaun Alexander with 112 touchdowns as illustrated by the bar chart, where his bar is the longest and the numerical value of 112 is the highest among all the listed players.\\
\textbf{answer2.name:} Assistant B\\
\textbf{answer2.answer}: The answer is Shaun Alexander. Shaun Alexander has 112 career touchdowns, the most of any player on the Seattle Seahawks. Steve Largent is second with 101 career touchdowns.\\
\vspace{2pt}\\
\textbf{result1-1.name}: Judge VLM 1\\
\colorbox{orange!30}{\textbf{result1-1.judge}: C}\\
\textbf{result1-1.analysis}: Both assistants correctly answer the question and provide the same answer with sufficient supporting details.\\
\textbf{result1-2.name}: Judge VLM 2\\
\colorbox{orange!30}{\textbf{result1-2.judge}: B}\\
\textbf{result1-2.analysis}: Both Assistant A and Assistant B correctly identified Shaun Alexander as the career touchdown leader for the Seattle Seahawks with 112 touchdowns, as shown in the bar chart. Assistant B provided additional context by mentioning that Steve Largent is second with 101 career touchdowns, offering a comparative perspective. While Assistant A's answer is accurate, Assistant B's answer adds a layer of detail that enriches the response without detracting from its relevance or accuracy.
\end{tcolorbox}

\begin{tcolorbox}[colback=gray!00,
                  colframe=black,
                  width=17.2cm,
                  arc=1.5mm, auto outer arc,
                  breakable,
                  left=0.9mm, right=0.9mm,
                  boxrule=0.9pt, colbacktitle = black!65!black,
                  title = {An example of batch ranking}
                 ]
    \begin{center}
    \includegraphics[width = 0.4\linewidth]{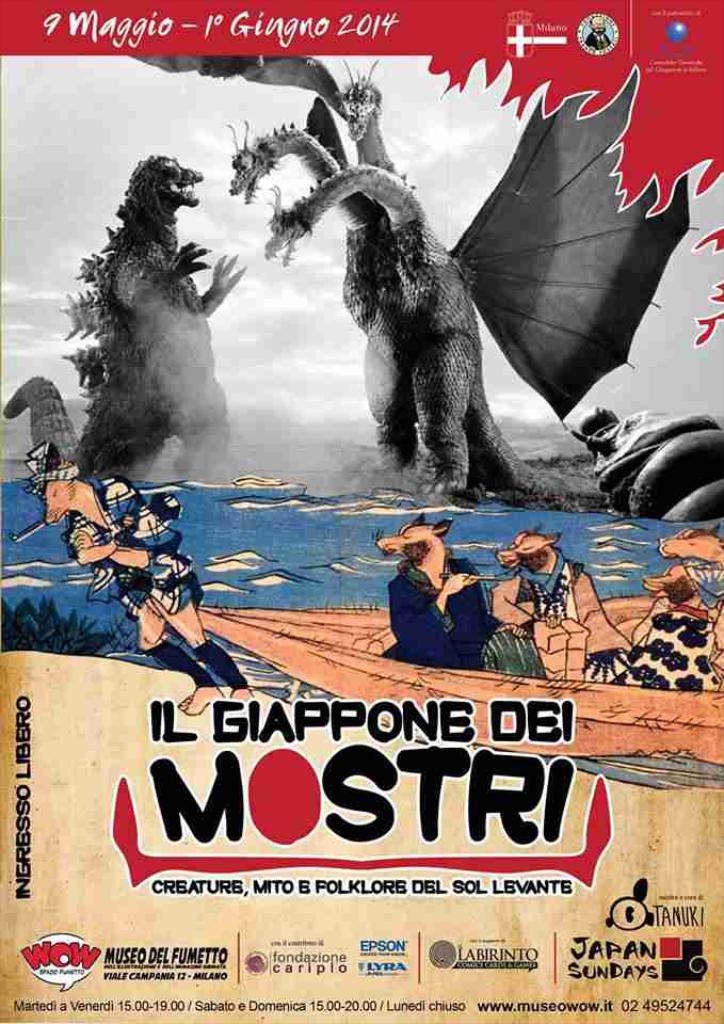}
    \end{center}
\colorbox{orange!30}{\textbf{Instruction:} This is a task of text reading on natural image. Please analyze this figure in detail and answer the } \colorbox{orange!30}{following question with reason based on this figure. what year is at the top of the magazine?} \\
\vspace{2pt}\\
\textbf{answer1.name:} Assistant A\\
\textbf{answer1.answer}: The year at the top of the magazine is 2014.\\
\textbf{answer2.name:} Assistant B\\
\textbf{answer2.answer}: The year at the top of the magazine is 2014. It is prominently displayed in large, bold letters, making it easily readable and the focal point of the image.\\
\textbf{answer3.name:} Assistant C\\
\textbf{answer3.answer}: The year at the top of the magazine is 2014. This can be determined by reading the text that says '9 Maggio – 10 Giugno 2014,' which indicates that the magazine or event was from May 9 to June 10, 2014. This text is found in the upper section of the image, in a red-colored font, just above the image of the two monsters.\\
\textbf{answer4.name:} Assistant D\\
\textbf{answer4.answer}: The year at the top of the magazine is 2014. The top of the magazine says ``9 MAGGIO - 1 GIUGNO 2014". ``9 MAGGIO - 1 GIUGNO" means ``May 9 - June 1". ``2014" is the year.\\
\vspace{2pt}\\
\textbf{result1-1.name}: Judge VLM 1\\
\textbf{result1-1.analysis}:
Assistant A: The assistant provides a concise and straightforward response, stating that the year at the top of the magazine is 2014. However, the response lacks any further explanation or supporting details. 
Assistant B: Similar to Assistant A, this assistant also briefly states that the year is 2014, but it adds a bit more detail by mentioning that the year is displayed prominently in large, bold letters, making it easily readable and the focal point of the image.
Assistant C: Assistant C offers a more comprehensive response by not only stating the year as 2014 but also providing a clear explanation. The assistant points out the specific text on the magazine that indicates the date range of May 9 to June 10, 2014, and correctly concludes that the year at the top is 2014.
Assistant D: Assistant D's response is comparable to that of Assistant C, as it includes both the identification of the year as 2014 and an explanation. However, Assistant D's explanation is slightly different, as it mentions the Italian translation of the date range, which is 9 MAGGIO - 1 GIUGNO 2014. The response also includes additional information about the event, mentioning that it took place in Milan and was organized by WOW Spazio Fumetto.

\colorbox{orange!30}{Judgement}: Based on the analysis, \colorbox{orange!30}{I rank the responses as follows: [[C]], [[D]], [[B]], [[A]]} Assistant C provides the most comprehensive and detailed response, clearly addressing the user's instruction and inquiry by identifying the year and providing supporting evidence from the image. Assistant D also offers a thorough explanation, along with additional information about the event. Assistant B's response is concise but lacks the level of detail provided by Assistants C and D. Assistant A's response is the briefest and lacks any explanation or supporting information. \\
\textbf{result1-2.name}: Judge VLM 2\\
\textbf{result1-2.analysis}: Assistant A: The response correctly identifies the year at the top of the magazine as 2014. However, it provides no additional information or context to support the answer., Assistant B: This response also correctly identifies the year as 2014 and goes further to describe the appearance of the text, which supports the user's instruction for detail and provides reasoning for the answer., Assistant C: The response is accurate in identifying the year as 2014 and provides a detailed description of where the date is located and what the text around it indicates. It adds context by mentioning the text color and the image below the date, giving a more comprehensive understanding., Assistant D: Assistant D accurately identifies the year as 2014 and gives a translation of the other text around the year, providing additional context. However, it slightly misquotes the text, stating '1 GIUGNO' instead of `10 Giugno,' which could be seen as a minor error in accurately reading the text. \colorbox{orange!30}{Judgement: Rank: [[C], [B], [A], [D]]}
\end{tcolorbox}


\end{document}